\documentclass[lettersize,journal]{IEEEtran}
\usepackage{amsmath,amsfonts}
\usepackage{algorithmic}
\usepackage{algorithm}
\usepackage{array}
\usepackage[caption=false,font=normalsize,labelfont=sf,textfont=sf]{subfig}
\usepackage{tikz}
\usepackage{amsmath}
\usepackage{booktabs}
\usepackage{adjustbox}
\usepackage{multirow}
\usepackage{epstopdf}
\usepackage{enumitem}
\usepackage{mathrsfs} 
\usepackage{enumitem}
\usepackage{mathrsfs} 
\usepackage{bm} 
\usepackage{amsfonts}
\usepackage{color}
\usepackage{xspace}
\usepackage{url}
\usepackage{multirow}
\usepackage{enumitem}
\usepackage{mathrsfs} 
\usepackage{bm} 
\usepackage{amsfonts}
\usepackage{color}
\usepackage{xspace}
\usepackage{url}
\usepackage{multirow}
\usepackage{adjustbox}
\usepackage{tikz}
\usepackage{amsmath}
\usepackage{epsfig}
\usepackage{graphicx}
\usepackage{amsmath}
\usepackage{algorithm}
\usepackage{algorithmic}
\usepackage{tcolorbox}
\usepackage{lipsum} 
\usepackage{filecontents}
\usepackage{verbatim}
\newcommand{\sys}{{\sc Armor}\xspace}
\usepackage{colortbl}

\usepackage{color}
\usepackage{xspace}
\usepackage{url}
\usepackage{multirow}
\usepackage{adjustbox}
\usepackage{tikz}
\usepackage{amsmath}
\usepackage{epsfig}
\usepackage{graphicx}
\usepackage{amsmath}
\usepackage{algorithm}
\usepackage{algorithmic}
\usepackage{soul}
\usepackage{filecontents}
\usepackage{verbatim}
\usepackage{multirow}
\usepackage{threeparttable}

\usepackage{booktabs}

\begin{document}

\title{\sys: Shielding Unlearnable Examples against Data Augmentation}

\author{
    
        Xueluan Gong, ~\IEEEmembership{Member, ~IEEE}, Yuji Wang,
        Yanjiao Chen,~\IEEEmembership{Senior Member,~IEEE,} Haocheng Dong, Yiming Li, Mengyuan Sun, Shuaike Li, Qian Wang,~\IEEEmembership{Fellow,~IEEE}, and Chen Chen
       
\IEEEcompsocitemizethanks{

\IEEEcompsocthanksitem X. Gong and C. Chen are with Nanyang Technological University, Singapore (E-mail: \{xueluan.gong, chen.chen\}@ntu.edu.sg).
\IEEEcompsocthanksitem Y. Chen is with the College of Electrical Engineering, Zhejiang University, China (E-mail: chenyanjiao@zju.edu.cn).
\IEEEcompsocthanksitem Y. Wang, H. Dong, M. Sun, S. Li, and Q. Wang are with the School of Cyber Science and Engineering, Wuhan University, China (E-mail:\{yujiwang, hc\_dong, mengyuansun, shuaikeli, qianwang\}@whu.edu.cn).
\IEEEcompsocthanksitem Y. Li is with the ZJU-Hangzhou Global Scientific and Technological Innovation Center (HIC) and also with the State Key
Laboratory of Blockchain and Data Security, Zhejiang University, China (e-mail: li-ym@zju.edu.cn).
%

}
}


\maketitle

\begin{abstract}
Private data, when published online, may be collected by unauthorized parties to train deep neural networks (DNNs). To protect privacy, defensive noises can be added to original samples to degrade their learnability by DNNs. Recently, unlearnable examples \cite{huang2021unlearnable} are proposed to minimize the training loss such that the model learns almost nothing. However, raw data are often pre-processed before being used for training, which may restore the private information of protected data. In this paper, we reveal the data privacy violation induced by data augmentation, a commonly used data pre-processing technique to improve model generalization capability, which is the first of its kind as far as we are concerned. We demonstrate that data augmentation can significantly raise the accuracy of the model trained on unlearnable examples from 21.3\% to 66.1\%. To address this issue, we propose a defense framework, dubbed \sys, to protect data privacy from potential breaches of data augmentation. To overcome the difficulty of having no access to the model training process, we design a non-local module-assisted surrogate model that better captures the effect of data augmentation. In addition, we design a surrogate augmentation selection strategy that maximizes distribution alignment between augmented and non-augmented samples, to choose the optimal augmentation strategy for each class. We also use a dynamic step size adjustment algorithm to enhance the defensive noise generation process. Extensive experiments are conducted on 4 datasets and 5 data augmentation methods to verify the performance of \sys. Comparisons with 6 state-of-the-art defense methods have demonstrated that \sys can preserve the unlearnability of protected private data under data augmentation. \sys reduces the test accuracy of the model trained on augmented protected samples by as much as 60\% more than baselines. We also show that \sys is robust to adversarial training. We will open-source our codes upon publication.
\end{abstract}


\section{Introduction}

\IEEEPARstart{T}{he} success of deep learning models is, to a great extent, attributed to the availability of expansive data samples \cite{najafabadi2015deep,wang2020recent,sun2017revisiting}. To train a well-performed DNN model, the model trainer may collect data samples from various sources. Although many public datasets can be easily downloaded online, private data samples are enticing thanks to their large quantity and diversity. For instance, an enormous number of facial images are published on social media and accessible to unauthorized parties to train well-performed face recognition DNN models. 
To protect personal data from being abused, worldwide regulations have been enacted, such as the European Union's General Data Protection Regulation (GDPR) \cite{voigt2017eu} and California Consumer Privacy Act (CCPA) \cite{goldman2020introduction}.

To protect private data samples from being used as training data for DNN models, researchers have proposed to add defensive noises to original samples. The resulting protected samples can be published as they have little learnability but maintain the perceptual quality of original samples. To achieve this objective, early works borrow techniques from adversarial example attacks, creating imperceptible adversarial noises to mislead the training process of DNN models \cite{madry2017towards, fowl2021adversarial, yuan2021neural}. Recently, the concept of unlearnable examples has been proposed \cite{huang2021unlearnable, liu2021going}. The main rationale is to minimize the training loss such that little can be learned from protected samples. Uncovering that adversarial training undermines unlearnability, Fu et al. \cite{fu2022robust} proposed to generate robust unlearnable examples that minimize adversarial training loss. 

In this paper, we present a new threat to data privacy posed by data augmentation, a commonly used data pre-processing technique \cite{cubuk2020randaugment,rebuffi2021data}. Data augmentation is usually performed on raw data samples to resolve class imbalance issues and improve the generalization capability of the trained model. As far as we are concerned, we are the first to unveil the potential data privacy threat induced by data augmentation. We demonstrate that state-of-the-art unlearnable examples (e.g., EMIN \cite{huang2021unlearnable}) can effectively suppress the accuracy of trained models to no more than 15\%. Unfortunately, if data augmentation is applied to unlearnable examples, the accuracy of trained models will be elevated to more than 60\%, severely damaging the unlearnability of protected data. 

To address this challenge, we develop \sys, a defense framework that can protect data privacy against advanced data augmentation. The main challenge is that the defender has no knowledge or control over the training process conducted by the attacker. More specifically, the defender does not know the model structure and the data augmentation strategy chosen by the attacker. To overcome this difficulty, we propose to use a carefully designed surrogate model and a surrogate augmentation selection strategy for generating defensive noises that minimize the training loss. For surrogate model construction, we design a non-local module that widens the receptive field of the surrogate model to capture more benefits from data augmentation. For surrogate augmentation strategy selection, we search for the best augmentation strategy to ensure that the augmented samples follow the distribution of the original samples. Under such strong surrogate models and surrogate strategies, the defensive noises are optimized via gradient descent. To further enhance the optimization process, we adopt a learning rate scheduling mechanism to dynamically adjust the learning rate based on the gradient norm. 

We have conducted extensive experiments to evaluate the data protection performance of \sys. We compare \sys with 6 state-of-the-art defense methods, including EMAX \cite{madry2017towards}, TAP \cite{fowl2021adversarial}, NTGA \cite{yuan2021neural}, EMIN \cite{huang2021unlearnable}, REM \cite{fu2022robust}, and random noise, on four datasets, i.e., CIFAR-10, CIFAR-100, Mini-ImageNet, and VGG-Face. The experiment results confirm that \sys can reduce the test accuracy of the model trained on the augmented protected dataset by as much as 60\% more than baselines. Furthermore, \sys showcases resilience against sophisticated data augmentation strategies, including Mixup \cite{zhang2017mixup}, Feature distillation \cite{liu2019feature}, PuzzleMix \cite{kim2020puzzle}, FastAA \cite{lim2019fast}, and DeepAA \cite{zheng2022deep}. It is also verified that \sys is robust to adversarial training.

To conclude, we make the following contributions.

\begin{itemize}
      \item We reveal the potential data privacy violation caused by data augmentation. To the best of our knowledge, we are the first to discover such a potential data privacy threat. 

      \item We develop a defense framework that can effectively resist advanced data augmentation strategies. To render the defense effective in the black-box setting, we carefully build the surrogate model and select the surrogate augmentation to generate robust defensive noises. 
            
      \item Extensive experiments show that \sys surpasses state-of-the-art data protection methods by significantly reducing model test accuracy under advanced data augmentation strategies. Moreover, \sys exhibits robustness against adversarial training and proves to be effective in generating class-wise defensive noises.
\end{itemize}

\section{Preliminaries}\label{Background}

\begin{table*}[tt]
	\caption{Test accuracy of models trained on protected samples with or without data augmentation. The augmentation method is DeepAA \cite{zheng2022deep}. Protected samples are generated by Gaussian Noise, EMIN \cite{huang2021unlearnable}, and REM \cite{fu2022robust}. } 
	\label{tab:pre-1}
	\centering
	\begin{tabular}{cccccccc}
		\toprule
 \multirow{9}{*}{CIFAR-10} & & & VGG-16 & ResNet-18 & ResNet-50& DenseNet-121& WRN$\_$34$\_$10 \\
 & \multirow{2}{*}{No protection} & No augmentation & 92.66\% & 94.09\% & 94.38\% & 94.89\% & 95.52\%\\
 & & Augmentation &  93.86\% & 94.49\% & 94.52\% & 95.28\% & 96.54\%\\
  \cline{2-8}
 & \multirow{2}{*}{Gaussian} & No augmentation & 92.49\% &  94.11\% & 93.56\% & 94.76\% & 95.61\%\\
 &  & Augmentation & 93.68\% & 92.79\%& 90.80\% & 93.63\%& 96.17\%\\
  \cline{2-8}
&   \multirow{2}{*}{EMIN \cite{huang2021unlearnable}} & No augmentation & 25.59\% & 25.08\% & 19.19\% & 21.70\% & 21.38\% \\
  & & Augmentation & \cellcolor[gray]{0.8}62.69\% &\cellcolor[gray]{0.8} 56.05\%  & \cellcolor[gray]{0.8}51.01\% & \cellcolor[gray]{0.8}57.32\% & \cellcolor[gray]{0.8}66.14\% \\
   \cline{2-8}
  & \multirow{2}{*}{REM \cite{fu2022robust}} & No augmentation & 29.61\% & 24.15\% & 20.50\% & 23.55\% & 24.50\% \\
&   & Augmentation & \cellcolor[gray]{0.8}47.58\% & \cellcolor[gray]{0.8}43.25\%  & \cellcolor[gray]{0.8}41.24\% & \cellcolor[gray]{0.8}44.39\% & \cellcolor[gray]{0.8}50.38\% \\
   \midrule
        
		 \multirow{9}{*}{CIFAR-100} && & VGG-16 & ResNet-18 & ResNet-50& DenseNet-121& WRN$\_$34$\_$10 \\
 & \multirow{2}{*}{No protection} & No augmentation & 69.43\%  & 71.84\% & 71.69\% & 73.04\% & 75.77\%\\
 & & Augmentation &  65.06\% & 75.01\% & 73.42\% & 76.06\% & 79.25\%\\
\cline{2-8}
 &  \multirow{2}{*}{Gaussian} & No augmentation & 69.30\% &  71.22\% & 69.61\% & 72.67\% & 75.05\%\\
  & & Augmentation & 64.04\% & 72.15\%& 72.62\% & 73.76\%& 76.92\%\\
 \cline{2-8}
 &  \multirow{2}{*}{EMIN \cite{huang2021unlearnable}} & No augmentation & 10.81\% & 15.19\% & 11.24\% & 13.64\% & 10.95\% \\
&   & Augmentation & \cellcolor[gray]{0.8}40.36\% &\cellcolor[gray]{0.8} 39.92\%  & \cellcolor[gray]{0.8}42.45\% & \cellcolor[gray]{0.8}40.32\% & \cellcolor[gray]{0.8}46.03\% \\
\cline{2-8}
  & \multirow{2}{*}{REM \cite{fu2022robust}} & No augmentation & 15.10\% & 13.18\% & 11.91\% & 13.37\% & 11.90\% \\
  & & Augmentation & \cellcolor[gray]{0.8} 30.70\% & \cellcolor[gray]{0.8}27.07\%  & \cellcolor[gray]{0.8}25.74\% & \cellcolor[gray]{0.8}32.74\% & \cellcolor[gray]{0.8}30.25\% \\
		\midrule
  		 \multirow{9}{*}{Mini-ImageNet} && & VGG-16 & ResNet-18 & ResNet-50& DenseNet-121& WRN$\_$34$\_$10 \\
 & \multirow{2}{*}{No protection} & No augmentation & 75.69\%  & 74.56\% & 68.34\% & 76.07\% & 71.11\%\\
 & & Augmentation &  77.66\% & 77.87\% & 70.50\% & 72.53\% & 74.30\%\\
\cline{2-8}
 &  \multirow{2}{*}{Gaussian} & No augmentation & 59.37\% &  65.55\% & 59.60\% & 68.45\% & 60.69\%\\
  & & Augmentation & 67.90\% & 70.84\%& 69.35\% & 68.64\%& 68.56\%\\
 \cline{2-8}
 &  \multirow{2}{*}{EMIN \cite{huang2021unlearnable}} & No augmentation & 14.98\% & 11.67\% & 16.81\% & 15.45\% & 22.96\% \\
&   & Augmentation & \cellcolor[gray]{0.8}58.06\% &\cellcolor[gray]{0.8} 63.51\%  & \cellcolor[gray]{0.8}58.84\% & \cellcolor[gray]{0.8}53.08\% & \cellcolor[gray]{0.8}50.38\% \\
\cline{2-8}
  & \multirow{2}{*}{REM \cite{fu2022robust}} & No augmentation & 26.81\% & 23.43\% & 24.02\% & 33.97\% & 33.40\% \\
  & & Augmentation & \cellcolor[gray]{0.8} 44.46\% & \cellcolor[gray]{0.8}43.60\%  & \cellcolor[gray]{0.8}42.52\% & \cellcolor[gray]{0.8}53.95\% & \cellcolor[gray]{0.8}47.92\% \\
		\midrule
   		 \multirow{9}{*}{VGG-Face} && & VGG-16 & ResNet-18 & ResNet-50& DenseNet-121& WRN$\_$34$\_$10 \\
 & \multirow{2}{*}{No protection} & No augmentation & 93.59\%  & 93.09\% & 93.62\% & 95.56\% & 94.54\%\\
 & & Augmentation &  97.18\% & 97.21\% & 93.02\% & 97.29\% & 97.73\%\\
\cline{2-8}
 &  \multirow{2}{*}{Gaussian} & No augmentation & 94.24\% &  94.28\% & 95.80\% & 96.20\% & 93.65\%\\
  & & Augmentation & 97.21\% & 97.61\%& 95.78\% & 90.91\%& 95.12\%\\
 \cline{2-8}
 &  \multirow{2}{*}{EMIN \cite{huang2021unlearnable}} & No augmentation & 1.46\% & 1.49\% & 1.57\% & 1.53\% & 1.65\% \\
&   & Augmentation & \cellcolor[gray]{0.8}42.63\% &\cellcolor[gray]{0.8} 48.75\%  & \cellcolor[gray]{0.8}41.14\% & \cellcolor[gray]{0.8}40.84\% & \cellcolor[gray]{0.8}35.75\% \\
\cline{2-8}
  & \multirow{2}{*}{REM \cite{fu2022robust}} & No augmentation & 3.21\% & 4.32\% & 5.53\% & 3.46\% & 4.21\% \\
  & & Augmentation & \cellcolor[gray]{0.8} 36.86\% & \cellcolor[gray]{0.8}46.43\%  & \cellcolor[gray]{0.8}40.78\% & \cellcolor[gray]{0.8}42.76\% & \cellcolor[gray]{0.8}37.43\% \\
  \bottomrule
	\end{tabular}
\end{table*}

\subsection{Data Privacy in Deep Learning}
A deep neural network\footnote{Without loss of generality, we consider classification tasks in this paper and will explore the issue in generative DNNs in our future works.} (DNN) is a function $f_\Theta$ parameterized by $\Theta$, mapping an input $\mathbf{x}$ to the output $\mathbf{y}$. The parameters $\Theta$ are optimized by minimizing the aggregated prediction error (i.e., defined as the loss function $\mathcal{L}$) on the training dataset $\mathcal{D}_{train} = \{\mathbf{x}_i,\mathbf{y}_i\}_{i=1}^N$ as
\begin{equation}
\min_{\Theta} \frac{1}{N} \sum_{i=1}^N \mathcal{L}(f_\Theta(\mathbf{x}_i), \mathbf{y}_i).
\end{equation}
 
Training datasets are considered as valuable assets since their quality has a considerable influence on the model performance. Training data samples are usually collected from various sources, including individual users whose private data may be sensitive \cite{koti2021swift,al2019privacy}. Potential data privacy breaches may occur in the pre-deployment and the post-deployment phases of deep learning \cite{gong2022private,shan2020fawkes}. In the pre-deployment phase, attackers may collect a large pool of unauthorized data (e.g., from online social networks) to construct their training datasets \cite{shan2020fawkes,fowl2021preventing,radiya2021data,hu2022protecting,cherepanova2021lowkey,huang2021unlearnable,fu2022robust}. In the post-deployment phase, attackers may infer the membership \cite{shokri2017membership,carlini2022membership,tang2022mitigating} or attributes \cite{gong2018attribute, fredrikson2015model,gong2023netguard} of training data samples. 

In this paper, we focus on protecting data privacy in the pre-deployment phase. More specifically, the main objective is to protect private user data from being exploited by unauthorized parties to train a well-performed DNN model. To attain this goal, existing research works have proposed to add defensive noises to original data samples to reduce their learnability, i.e., the DNN model trained on the perturbed data samples will not reach satisfactory accuracy. Given an original sample $\mathbf{x}$, its privacy-preserving version $\mathbf{x}'$ is created as
\begin{equation}
    \mathbf{x}^{\prime} = \mathbf{x} + \mathbf{\delta}.
\end{equation}
where $\delta$ is the defensive noise. 

There are various choices for $\delta$, a naive one being the Gaussian noise. Adversarial noises are often adopted as they may mislead the training process. EMAX \cite{madry2017towards}, UTAP \cite{fowl2021adversarial}, CTAP \cite{fowl2021adversarial} and NTGA \cite{yuan2021neural} all followed this line. Recently, the idea of unlearnable samples emerged \cite{huang2021unlearnable}, generating the defensive noise $\delta$ via an optimization problem
\begin{equation}\label{Emin-equa}
\min_{\theta} \mathbb{E}_{(\mathbf{x}, \mathbf{y}) \in \mathcal{D}_c} [\min_{\mathbf{\delta}} \mathcal{L}(f_{\theta}(\mathbf{x} + \mathbf{\delta}), \mathbf{y})],
\end{equation}
where $f_{\theta}$ represents a hypothetical DNN model under training, $\mathcal{D}_c$ is a clean training dataset used to train $f_{\theta}$, and $\mathcal{L}$ is the loss function for training $f_{\theta}$. The bi-level optimization problem consists of an outer and an inner minimization problems. The outer minimization problem searches for the parameters $\theta$ that minimize the training error of $f_{\theta}$ on clean training dataset $\mathcal{D}_c$. The inner minimization problem searches for an $L_p$-norm bounded noise $\delta$ that minimizes the loss of $f_{\theta}$ on sample $\mathbf{x}+\mathbf{\delta}$ such that model $f_{\theta}$ learns almost nothing from sample  $\mathbf{x}+\mathbf{\delta}$.

Despite the effort to minimize the learnability of original samples, unlearnable examples are found to be vulnerable to adversarial training \cite{fu2022robust}. More specifically, adversarial examples created based on these unlearnable examples will yield non-minimal loss, allowing the model to learn useful information, thus undermining privacy protection. To tackle this issue, Fu et al. \cite{fu2022robust} proposed a robust unlearnable example generation method using min-min-max optimization as 
\begin{equation}\label{REM-equa}
\min_\theta \mathbb{E}_{(\mathbf{x},\mathbf{y}) \in \mathcal{D}_c} \min_{\delta} \max_{\epsilon} \mathcal{L}(f_\theta(\mathbf{x}+\epsilon + \delta), \mathbf{y}),
\end{equation}
where $\epsilon$ is the adversarial noise added to maximize the loss for adversarial training. Both $\epsilon$ and $\delta$ are bounded to satisfy imperceptibility. 

In this paper, we will reveal the vulnerability of existing defense methods, especially unlearnable examples, to data augmentation. We further propose corresponding countermeasures to strengthen data privacy protection.

\subsection{Data Augmentation}
Data augmentation \cite{van2001art,maharana2022review,tanner1987calculation} is a technique that enriches a training dataset by generating new examples through transformations of original data samples, e.g., flipping, rotating, scaling, cropping, and changing color. The purpose of data augmentation is to help improve the performance and generalization capability of machine learning models, especially if the available training data samples are limited. By generating more diverse examples, data augmentation helps the model learn to be more robust to data variations. It is shown that data augmentation can also help defend against adversarial example attacks \cite{qiu2020fencebox,zeng2020data} and backdoor attacks \cite{borgnia2021strong,qiu2021deepsweep}. In this paper, we review five state-of-the-art representative data augmentation methods, i.e., Mixup \cite{zhang2017mixup}, feature distillation \cite{liu2019feature}, PuzzleMix \cite{kim2020puzzle}, Fast AutoAugment \cite{lim2019fast}, and DeepAA \cite{zheng2022deep}.

\emph{\underline{Mixup.}}
Mixup \cite{zhang2017mixup} constructs a weighted combination of randomly selected pairs of samples from the training dataset. Given two samples $(\mathbf{x}_i, \mathbf{y}_i)$ and $(\mathbf{x}_j, \mathbf{y}_j)$, an augmented sample is generated as 
$(\lambda \mathbf{x}_i + (1-\lambda) \mathbf{x}_j, \lambda \mathbf{y}_i + (1-\lambda) \mathbf{y}_j)$, where $\lambda \in [0,1]$ is a mixing coefficient that determines the weight of each sample in the linear combination. 

\emph{\underline{Feature distillation.}}
Feature distillation \cite{liu2019feature} was initially designed to mitigate adversarial example attacks. It redesigns the quantization procedure of a traditional JPEG compression algorithm \cite{wallace1992jpeg}. Depending on the location of the derived quantization in JPEG, feature distillation has the one-pass and the two-pass modes. The one-pass feature distillation inserts a quantization/de-quantization in the decompression process of the original JPEG algorithm. The two-pass feature distillation embeds a crafted quantization at the sensor side to compress the raw data samples. 

\emph{\underline{PuzzleMix.}}
PuzzleMix \cite{kim2020puzzle} is an extension of Mixup, resolving the problem of generating unnatural augmented samples. PuzzleMix finds the optimal mixing mask based on the saliency information. PuzzleMix is shown to have outperformed state-of-the-art Mixup methods in terms of generalization and robustness against data corruption and adversarial perturbations.

\emph{\underline{FastAA.}} AutoAugment \cite{cubuk2019autoaugment} automates the search for an optimal data augmentation strategy given the original dataset using reinforcement learning. To reduce the computational complexity of AutoAugment, Fast AutoAugment (FastAA) \cite{lim2019fast} uses a more efficient search strategy based on density matching. 
The main idea is to learn a probability distribution over a space of candidate augmentation strategy that maximizes the performance of a given model on a validation dataset. The distribution is learned by matching the density of a learned feature space of augmented samples to that of original samples. In this way, FastAA significantly speeds up the search process compared to the original AutoAugment algorithm.

\emph{\underline{DeepAA.}}
Deep AutoAugment (DeepAA) \cite{zheng2022deep} constructs a multi-layer data augmentation framework. In each layer, the augmentation strategy is optimized to maximize the cosine similarity between the gradient of the original data and that of the augmented data along the low-variance direction. To avoid an exponential increase in the dimensionality of the search space, DeepAA incrementally stacks layers according to the data distribution transformed by all prior augmentation layers.

Given a set of augmentation methods, the augmentation strategy of each layer is represented as a probability of using each augmentation method. 
\begin{equation}
\begin{split}\label{equ:cs}
\arg \max_{\mathbf{p}} \frac{\mathbf{g}^T \cdot \nabla \mathbf{g}^A(\mathbf{x}, \mathbf{p})}{||\mathbf{g}^T||\cdot||\nabla \mathbf{g}^A(\mathbf{x}, \mathbf{p})||},\\
\end{split}
\end{equation}
where $\mathbf{p}$ is the augmentation strategy, $\nabla \mathbf{g}^A(\mathbf{x}, \mathbf{p})$ is the average gradient of the augmented samples, $\mathbf{g}^T$ is the gradient of the original data samples, and $||\cdot||$ is the $L_2$-norm.

\section{Unveiling the Effects of Data Augmentation on Data Privacy}\label{sec:unveiling}
We reveal that data augmentation will impair the privacy protection provided by existing defensive noise generation methods. As far as we are concerned, this is the first time such a vulnerability has been exposed.

\begin{table*}[tt]
	\caption{{\color{black}Test accuracy of models trained on protected samples with or without data augmentation. Different data augmentation methods are adopted. Protected samples are generated by EMIN and REM. }}
	\label{tab:pre2}
	\centering
	\footnotesize
	\setlength\tabcolsep{8pt}
	\begin{tabular}{ccc|cccccccccccccc}
		\toprule
		&&&\multirow{1}{*}{\shortstack{CIFAR-10}}&
            \multirow{1}{*}{\shortstack{CIFAR-100}}
		&\multicolumn{1}{c}{\color{black}Mini-ImageNet}
          
		&\multicolumn{1}{c}{\color{black}VGG-Face}\\
        \multirow{5}{*}{\shortstack{\color{black}Mixup \cite{zhang2017mixup}}}
        &Unprotected&Augmentation&95.03\%&77.19\%&76.89\%&94.49\%\\
        \cline{2-7}
        & \multirow{2}{*}{EMIN \cite{huang2021unlearnable}}&No Augmentation&25.08\%&15.19\%&14.98\%&1.46\%\\
        &&Augmentation&\cellcolor[gray]{0.8}33.65\%&\cellcolor[gray]{0.8}18.02\%&\cellcolor[gray]{0.8}17.79\%&\cellcolor[gray]{0.8}47.96\%\\
        \cline{2-7}
        & \multirow{2}{*}{REM \cite{fu2022robust}}&No Augmentation&24.15\%&13.18\%&26.81\%&3.21\%\\
        &&Augmentation&\cellcolor[gray]{0.8}31.58\%&\cellcolor[gray]{0.8}17.61\%&\cellcolor[gray]{0.8}36.98\%&\cellcolor[gray]{0.8}36.76\%\\
        \midrule
        &&&\multirow{1}{*}{\shortstack{CIFAR-10}}&
            \multirow{1}{*}{\shortstack{CIFAR-100}}
		&\multicolumn{1}{c}{\color{black}Mini-ImageNet}
          
		&\multicolumn{2}{c}{\color{black}VGG-Face}\\
        \multirow{5}{*}{\shortstack{\color{black}Feature distillation \cite{liu2019feature}}}
        &Unprotected&Augmentation&92.91\%&68.48\%&77.11\%&95.40\%\\
        \cline{2-7}
        & \multirow{2}{*}{EMIN \cite{huang2021unlearnable}}&No Augmentation&25.08\%&15.19\%&14.98\%&1.46\%\\
        &&Augmentation&\cellcolor[gray]{0.8}31.17\%&\cellcolor[gray]{0.8}22.08\%&\cellcolor[gray]{0.8}21.40\%&\cellcolor[gray]{0.8}38.65\%\\
        \cline{2-7}
        & \multirow{2}{*}{REM \cite{fu2022robust}}&No Augmentation&24.15\%&13.18\%&26.81\%&3.21\%\\
        &&Augmentation&\cellcolor[gray]{0.8}32.67\%&\cellcolor[gray]{0.8}25.05\%&\cellcolor[gray]{0.8}36.13\%&\cellcolor[gray]{0.8}27.34\%\\
		\midrule
  &&&\multirow{1}{*}{\shortstack{CIFAR-10}}&
            \multirow{1}{*}{\shortstack{CIFAR-100}}
		&\multicolumn{1}{c}{\color{black}Mini-ImageNet}
          
		&\multicolumn{2}{c}{\color{black}VGG-Face}\\
    \multirow{5}{*}{\shortstack{\color{black}PuzzleMix \cite{kim2020puzzle}}}
        &Unprotected&Augmentation&95.37\%&68.65\%&76.46\%&94.36\%\\
        \cline{2-7}
        & \multirow{2}{*}{EMIN \cite{huang2021unlearnable}}&No Augmentation&25.08\%&15.19\%&14.98\%&1.46\%\\
        &&Augmentation&\cellcolor[gray]{0.8}40.52\%&\cellcolor[gray]{0.8}28.57\%&\cellcolor[gray]{0.8}18.14\%&\cellcolor[gray]{0.8}48.78\%\\
        \cline{2-7}
        & \multirow{2}{*}{REM \cite{fu2022robust}}&No Augmentation&24.15\%&13.18\%&26.81\%&3.21\%\\
        &&Augmentation&\cellcolor[gray]{0.8}40.43\%&\cellcolor[gray]{0.8}21.12\%&\cellcolor[gray]{0.8}64.95\%&\cellcolor[gray]{0.8}35.87\%\\
        \midrule
        &&&\multirow{1}{*}{\shortstack{CIFAR-10}}&
            \multirow{1}{*}{\shortstack{CIFAR-100}}
		&\multicolumn{1}{c}{\color{black}Mini-ImageNet}
          
		&\multicolumn{2}{c}{\color{black}VGG-Face}\\
        \multirow{5}{*}{\shortstack{\color{black}FastAA  \cite{lim2019fast}}}
        &Unprotected&Augmentation&94.81\%&73.64\%&76.60\%&94.97\%\\
        \cline{2-7}
        & \multirow{2}{*}{EMIN \cite{huang2021unlearnable}}&No Augmentation&25.08\%&15.19\%&14.98\%&1.46\%\\
        &&Augmentation&\cellcolor[gray]{0.8}44.00\%&\cellcolor[gray]{0.8}32.27\%&\cellcolor[gray]{0.8}18.94\%&\cellcolor[gray]{0.8}40.08\%\\
        \cline{2-7}
        & \multirow{2}{*}{REM \cite{fu2022robust}}&No Augmentation&24.15\%&13.18\%&26.81\%&3.21\%\\
        &&Augmentation&\cellcolor[gray]{0.8}38.40\%&\cellcolor[gray]{0.8}24.72\%&\cellcolor[gray]{0.8}38.73\%&\cellcolor[gray]{0.8}35.74\%\\
		\bottomrule
	\end{tabular}
\end{table*}
We first present the performance of different models trained on protected samples with or without data augmentation. As shown in Table \ref{tab:pre-1}, we assess the protection performance of Gaussian noise and two state-of-the-art defense methods based on unlearnable examples, i.e., EMIN \cite{huang2021unlearnable} and REM \cite{fu2022robust}. We use DeepAA \cite{zheng2022deep} as the data augmentation method.
 
 \begin{tcolorbox}[title = {Effect of Data Augmentation on Data Privacy}, colback=white]
{
$\bullet$ \textbf{Without data augmentation}, data privacy protection provided by unlearnable examples are effective. \\
$\bullet$ \textbf{With data augmentation},  data privacy protection provided by unlearnable examples are undermined.
}
\end{tcolorbox}

As shown in Table \ref{tab:pre-1}, without data augmentation, EMIN and REM effectively prevent DNN models from achieving high test accuracy by learning from protected data samples. For instance, DNNs trained on unprotected clean CIFAR-10 dataset yield test accuracies of 92.66\% (VGG-16) and 94.09\% (ResNet-18). These accuracies drop to 25.59\% (VGG-16) and 25.08\% (ResNet-18) under EMIN protection. Unfortunately, when data augmentation is adopted, EMIN and REM struggle to keep the test accuracy of unauthorized models below 30\%, especially for EMIN. For example, the test accuracy of the WRN$\_$34$\_$10 model is only 21.38\% under EMIN for CIFAR-10 dataset, but jumps to 66.14\% once DeepAA is utilized.

To further verify our findings, we assess other state-of-the-art data augmentation strategies, including Mixup \cite{zhang2017mixup}, Feature distillation \cite{liu2019feature}, PuzzleMix \cite{kim2020puzzle}, and FastAA \cite{lim2019fast}. The experimental results are presented in Table~\ref{tab:pre2}.
The results demonstrate that various data augmentation strategies can undermine data privacy protection. Besides DeepAA, both FastAA and PuzzMix can substantially improve model test accuracy (e.g., $>$60\% in several cases) even if EMIN and REM are used for privacy protection. More evaluation results of the effects of data augmentation on defense methods EMAX \cite{madry2017towards}, UTAP \cite{fowl2021adversarial}, CTAP \cite{fowl2021adversarial} and NTGA \cite{yuan2021neural} can be found in Table~\ref{tab:non-com} and Table~\ref{tab:com-with}.

Through comprehensive experiments, we have discovered that almost all existing defensive noises are susceptible to data augmentation. This finding highlights the importance of mitigating potential data privacy violations introduced by data augmentation. 

\begin{figure*}[tt]
    \centering
    \includegraphics[width=0.95\textwidth]{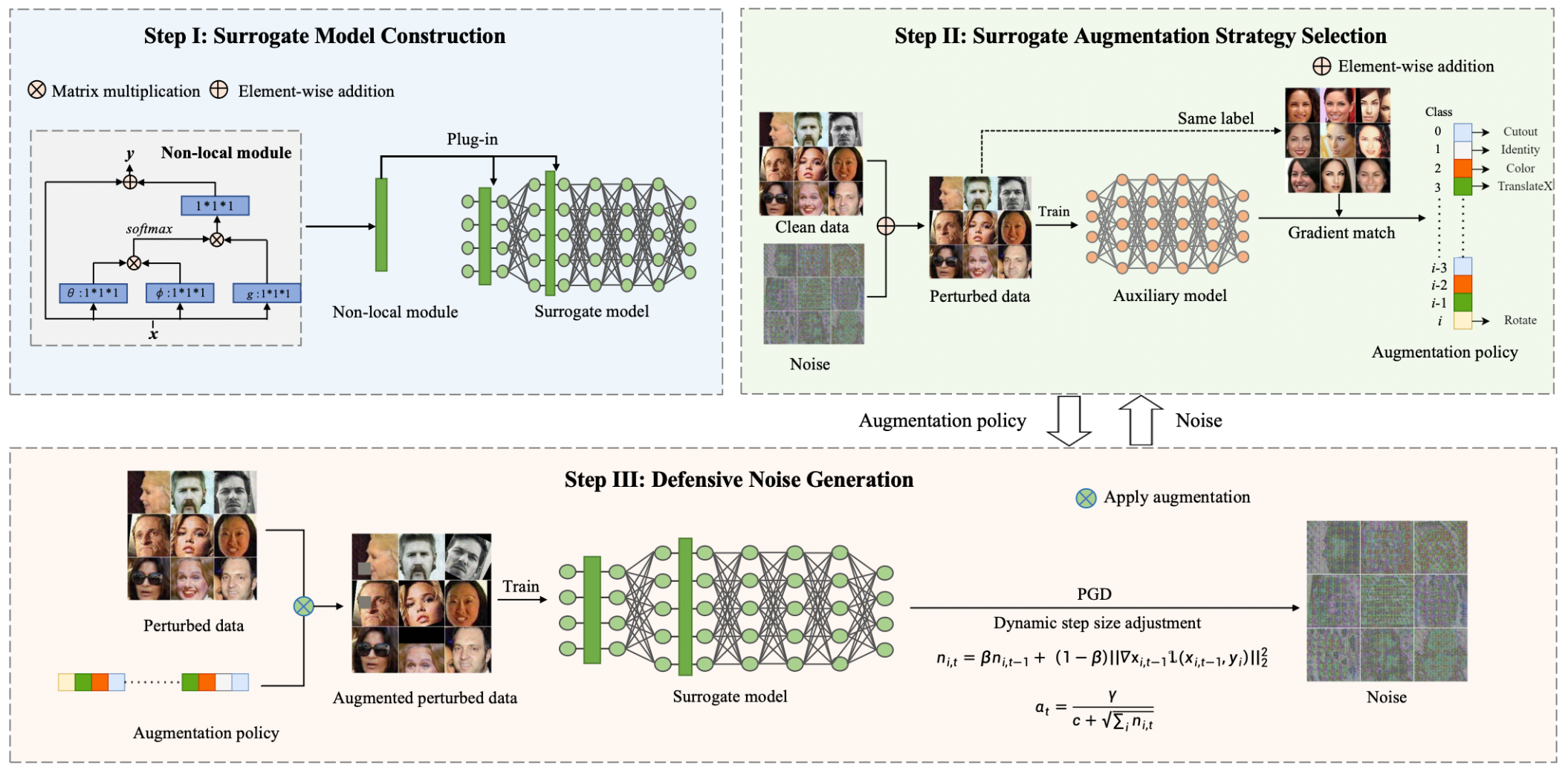}
    \caption{Overview of \sys. \sys mainly contains three key components, i.e., surrogate model construction, surrogate augmentation strategy selection, and defensive noise generation.}
    \label{overview}
\end{figure*}

\section{Protecting Data Privacy Under Data Augmentation}

In this paper, we propose \sys to enhance data privacy protection under the impact of data augmentation.

\subsection{Threat Model}
We consider two parties, i.e., the defender and the attacker. 

\textbf{Defender}. The role of the defender is usually assumed by a data owner who possesses a private dataset $\mathcal{D}_c={(\mathbf{x}_i,y_i)}_{i=1}^n$. The goal of the defender is to generate a corresponding protected dataset $\mathcal{D}_u$ to be published on social media, where $\mathcal{D}_u=\{(\mathbf{x}_i+\delta_i,y_i)\}_{i=1}^n$. 
It should be ensured that a model trained on the published protected dataset $\mathcal{D}_u$ performs poorly on the test set $\mathcal{D}_t$, even if the model trainer utilizes advanced data augmentation strategies. The defensive noise $\delta$ is bounded by $||\delta||_p \leq \epsilon$ to guarantee imperceptibility.
Following existing defense works \cite{huang2021unlearnable}, we assume the defender has full access to the private dataset $\mathcal{D}_c$. However, the defender does not have any knowledge or control over the training process of the attacker model. Additionally, the defender cannot modify the protected examples once they are published.

\textbf{Attacker}. 
The role of the attacker is usually assumed by a model trainer who aims to train a well-performed model $f$ with the public dataset $\mathcal{D}_u$. The attacker can employ any data augmentation strategy to improve the model performance. We also explore the effectiveness of \sys when the attacker adopts adversarial training to bolster model performance.

\subsection{Privacy Protection Problem Formulation}

To generate defensive noise $\delta$ that can resist data augmentation, we formulate the optimization problem as
\begin{equation}
\begin{split}
    & \min_{\theta}\mathbb{E}_{(\mathbf{x},\mathbf{y})\sim \mathcal{D}_c}[\min_\delta \mathcal{L}(f(\mathcal{A}(\mathbf{x}+\delta, \mathbf{p}),\mathbf{y})],\\
\end{split}
\label{optimization}
\end{equation}
where $\mathcal{A}(\mathbf{x}+\delta, \mathbf{p})$ is the function that enforces data augmentation strategy $\mathbf{p}$ on the protected sample $\mathbf{x}+\delta$. The key idea of Equation \eqref{optimization} is to minimize the loss of the protected sample even if data augmentation is applied. In Equation \eqref{optimization}, $f(\cdot)$ and $\mathbf{p}$ are the surrogate model and the surrogate augmentation strategy that the defender uses to mimic the real model and real augmentation strategy adopted by the attacker. Therefore, the choice of $f(\cdot)$ and $\mathbf{p}$ are essential to determine the privacy protection performance. In addition, to solve the min-min problem to attain the optimal defensive noise $\delta$ is non-trivia. To address these challenges, we equip \sys with three main building blocks, as depicted in Figure \ref{overview}. 

\begin{itemize}
\item \emph{Surrogate model construction.} { With no information on the structure of the attacker model, the defender may have to initialize a surrogate model. Our empirical study reveals that a surrogate model, established with the aid of data augmentation, can capture more informative features, leading to improved protection performance. Since data augmentation encourages the expansion of the model's receptive field to a broader sample region, we introduce a non-local module as a plug-in into the surrogate model to capture a global receptive field of the sample. }
\item \emph{Surrogate augmentation strategy selection.} Our extensive experiment results in Section \ref{sec:unveiling} have demonstrated that DeepAA \cite{zheng2022deep} raises the accuracy of the attack model the most. 
{Given an augmentation operation set, for each sample, DeepAA computes the optimal augmentation strategy to achieve gradient alignment, according to Equation \eqref{sec:unveiling}. Inspired by DeepAA, we propose an augmentation strategy selection algorithm to dynamically select class-specific augmentation strategies.}
\item \emph{Defensive noise generation.} We adopt the Projected Gradient Descent (PGD) \cite{madry2017towards} to solve the min-min optimization problem. PGD performs iterative updates to search for the optimal solution, {transforming the noise generation into an iterative process of augmentation policy update and noise update.} Besides, we propose to use an adaptive learning rate scheduling scheme in PGD, which enhances the efficiency of the iteration process and mitigates the risk of overfitting.

\end{itemize}

\subsection{Surrogate Model Construction}
Ideally, the surrogate model should have the same structure as the attacker model. However, we assume that the defender has no access to the attacker model. Previous works \cite{huang2021unlearnable, fu2022robust} have revealed that unlearnable examples generated by one specific model type can still show good protection performance under other trained model types. In this case, we may opt to choose a commonly used model structure, e.g., ResNet-18, for constructing the surrogate model.

Our extensive empirical study reveals that \sys provides better protection when the surrogate model benefits more from data augmentation. Based on this, we improve the fundamental surrogate model structure to amplify the impact of data augmentation.

Data augmentation aims to enhance model generalization capability by reducing the model's reliance on local features, which helps mitigate overfitting. By increasing data diversity, data augmentation encourages the model to enlarge the receptive field to a broader sample region. Inspired by this principle, after initializing the surrogate model with a commonly-used DNN model, we further incorporate a non-local module \cite{wang2018non} that captures a global receptive field of the sample. 

The non-local module transforms an input $\mathbf{a}$ into a same-dimensional output $\mathbf{b}$ as
\begin{equation}
    \mathbf{b}_{i}=\frac{1}{\mathcal{C}(\mathbf{a})} \sum_{\forall j} F\left(\mathbf{a}_{i}, \mathbf{a}_{j}\right) G\left(\mathbf{a}_{j}\right),
\end{equation}
where $\mathbf{b}_i$ is the $i$-th feature of output $\mathbf{b}$, $\mathbf{a}_i$ is the $i$-th feature of input $\mathbf{a}$, $\mathcal{F}(\cdot, \cdot)$ is a pairwise function whose output depends on the relationship between $\mathbf{a}_i$ and $\mathbf{a}_j$,  and $\mathcal{G}(\cdot)$ computes the $j$-th feature of the representation of input $\mathbf{a}$. In addition, $C(a)$ refers to a factor used for normalization.

The key to a non-local module is the design of $\mathcal{G}$ and $\mathcal{F}$. For $\mathcal{G}$, we choose a linear transformation as $\mathcal{G}(\mathbf{a}_j)=\mathbf{W}_G \mathbf{a}_j$, where $\mathbf{W}_G$ is a learned weight matrix obtained in the training process. For $\mathcal{F}$, we develop a Gaussian-embedded structure as
\begin{equation}
    \mathcal{F}\left(\mathbf{a}_{i}, \mathbf{a}_{j}\right)=e^{\varphi\left(\mathbf{a}_{i}\right)^{T} \cdot \phi\left(\mathbf{a}_{j}\right)}, 
\end{equation}
where $\varphi(\cdot)$ and $\phi(\cdot)$ are linear transformation functions, and $\mathbf{x}^T$ is the transportation of matrix $\mathbf{x}$.

The non-local module can be a plug-in to be inserted between any layers of a DNN model. 

\begin{table}[tt]
	\caption{A List of Standard Augmentation Search Space. ``Identity" denotes applying no data augmentation to the data sample.}
	\label{tab:auglist}
	\centering
 \setlength\tabcolsep{25pt}
	\footnotesize
\begin{tabular}{l|c}
\toprule
\text{Operation}                     & \text{Magnitude} \\ \hline
\multicolumn{1}{l|}{Identity} & -                  \\
\multicolumn{1}{l|}{ShearX}            & {[}-0.3, 0.3{]}    \\
\multicolumn{1}{l|}{ShearY}            & {[}-0.3, 0.3{]}    \\
\multicolumn{1}{l|}{TranslateX}        & {[}-0.45, 0.45{]}  \\
\multicolumn{1}{l|}{TranslateY}        & {[}-0.45, 0.45{]}  \\
\multicolumn{1}{l|}{Rotate}            & {[}-30, 30{]}      \\
\multicolumn{1}{l|}{AutoContrast}      & -                  \\
\multicolumn{1}{l|}{Invert}            & -                  \\
\multicolumn{1}{l|}{Equalize}          & -                  \\
\multicolumn{1}{l|}{Solarize}          & {[}0, 256{]}       \\
\multicolumn{1}{l|}{Posterize}         & {[}4, 8{]}         \\
\multicolumn{1}{l|}{Contrast}          & {[}0.1, 1.9{]}     \\
\multicolumn{1}{l|}{Color}             & {[}0.1, 1.9{]}     \\
\multicolumn{1}{l|}{Brightness}        & {[}0.1, 1.9{]}     \\
\multicolumn{1}{l|}{Sharpness}         & {[}0.1, 1.9{]}     \\
\multicolumn{1}{l|}{Flips}             & -                  \\
\multicolumn{1}{l|}{Cutout}            & {[}8, 16{]}        \\
\multicolumn{1}{l|}{Crop}              & -                  \\ \hline
\end{tabular}
\end{table}

\begin{algorithm}[tt]
\caption{Surrogate Augmentation Strategy Selection}
\begin{algorithmic}[1]\label{alg1}
    \REQUIRE Clean dataset $D_c$, existing noise $\delta$, existing augmentation strategy $\mathbf{p}$, pretrained auxiliary model $f^\prime$, num of classes $c$ and training epoch $r$.
    \ENSURE Augmentation strategy $\mathbf{p}$.
    
    \FOR{$k$ in $1, 2, ...r$}
    \STATE Sample a minibatch $(x_k,y_k)\sim D_c$. 
    \STATE $x'_k=x_k+\delta$.
    \STATE Update the auxiliary model $f^\prime$  based on minibatch $(x'_k,y_k)$.
    \ENDFOR
    \FOR{$i$ in $0, 1, 2, ...c - 1$}
    \STATE Randomly select sample batches $x_1$ and $x_2$ of the same label $i$. 
    \STATE //Gradient matching
    \STATE $\mathbf{p}_i= \arg\max_{o \in \mathbb{O}}\text{CS}(\nabla x_1,\nabla \mathcal{A}(x_2, \mathbf{o})),$ \\
    \ENDFOR
    
    \STATE Return $\mathbf{p}$.
\end{algorithmic}
\end{algorithm}
\subsection{Surrogate Augmentation Strategy Selection}

Data augmentation enriches a dataset by synthesizing additional data points that follow the same underlying distribution \cite{hataya2020faster}. Thus, an effective data augmentation should maintain the underlying data distribution, ensuring that the distribution of augmented samples aligns with that of the original samples \cite{chen2020group}. To achieve this goal, DeepAA adopts gradient alignment, i.e., aligning the gradients computed on the augmented sample batch with those computed based on the original sample batch \cite{zheng2022deep}. As the attacker only gets the protected version of private samples, we select the augmentation strategy that maximizes the cosine similarity between the average gradients of the protected data and those of the augmented protected data. 

Mathematically, we search for the optimal augmentation strategy $\mathbf{p}$ as 
\begin{equation}
\begin{split}
  \mathbf{p}=\arg\max_{\mathbf{p}}\text{CS}(\nabla x_1,\nabla \mathcal{A}(x_2, \mathbf{p}))\\
  =\arg\max_{\mathbf{p}} \frac{\nabla x_1 \times \nabla \mathcal{A}(x_2, \mathbf{p})}{||\nabla x_1||\times ||\nabla \mathcal{A} (x_2, \mathbf{p})||},
    \end{split}
\end{equation}
where $\text{CS}(\cdot,\cdot)$ calculates the cosine similarity, $||\cdot||$ represents the $L_2$ norm, and $x_1$ and $x_2$ are two batches of images from the current perturbed data. To improve efficiency, we adopt the class-wise augmentation strategy, i.e., we select a specific strategy for each class in every epoch, since samples from the same class usually share similar feature representations. 

For each class, we determine the optimal augmentation strategy using an auxiliary model $f^{\prime}$. This model is initially trained with clean samples and subsequently updated using perturbed data during each epoch of augmentation strategy selection. Let $\mathbf{p}_i$ represent the selected augmentation operation for the $i$-th class. The augmentation strategy selection is updated as 
\begin{equation}
    \mathbf{p}_{i}= \arg\max_{o \in \mathbb{O}}\text{CS}(\nabla x_1,\nabla \mathcal{A}(x_2, \mathbf{o}))\\,
\end{equation}
where $\mathbb{O}$ denotes the augmentation operation set as defined in Table~\ref{tab:auglist}, and $x_1$ and $x_2$ represent two batches of images labeled $i$. The algorithm for selecting the surrogate augmentation strategy is outlined in Algorithm~\ref{alg1}.

\begin{algorithm}[tt]
\caption{Defensive Noise Generation}
\begin{algorithmic}[1]\label{alg2}
    \REQUIRE Clean dataset $D_c$, existing noise $\delta$, existing augmentation strategy $\mathbf{p}$, training epoch of surrogate model $r_f$, and noise training epoch $r_d$.
    \ENSURE Updated noise $\delta$.
    \STATE // The process of noise update.

    \FOR{$k$ in $1, 2, ...r_f$}
    \STATE Sample a minibatch $(x_k,y_k)\sim D_c$. 
    \STATE Update surrogate model $f$ based on minibatch $(\mathcal{A}(x_k+\delta, \mathbf{p}),y_k)$.
    \ENDFOR 
    \STATE $\delta_{0}$ = $\delta$;
    \STATE //PGD method iteration with the adaptive step size; the first subscript indicates the image index, the second subscript indicates the number of update rounds; $\beta, c, \gamma$ are default constants.
    \FOR{$t$ in $1, 2, ...r_d$}
    \STATE $n_{i,t}=\beta n_{i,t-1} + (1-\beta)||\nabla_{x_{i,t-1}} \mathrm{L}(x_{i,t-1}, y_i)||_2^2$.
    \STATE $\alpha_t=\frac{\gamma}{c+\sqrt{\sum_i n_{i,t}}}$.
    \STATE $\delta_{t}= \delta_{t-1}-\alpha_t*\mathrm{sign}(\nabla_{\delta}\mathcal{L}(f(\mathcal{A}(\mathbf{x}+\delta_{t-1}, \mathbf{p}),\mathbf{y}))),$
    \ENDFOR
    
    \STATE Return $\delta$.
\end{algorithmic}
\end{algorithm}

\begin{figure*}[tt]
	\centering
	\hspace{-0cm}
	\begin{minipage}[t]{3.44in}
		\centering
		\includegraphics[trim=0mm 0mm 0mm 0mm, clip,width=3.4in]{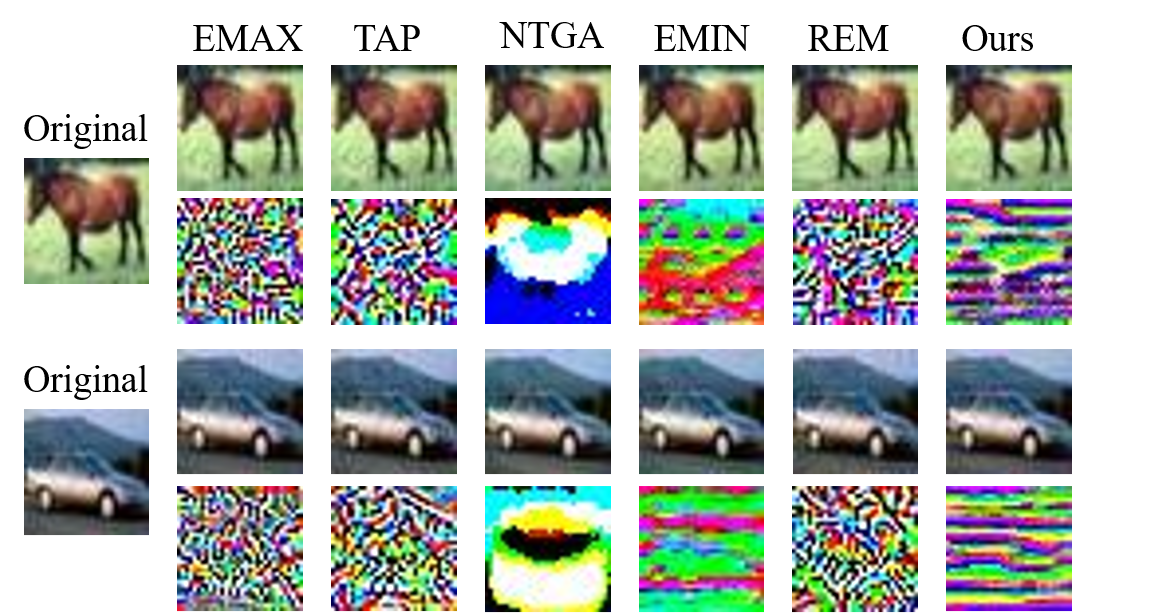}
		\centerline{\footnotesize (a) CIFAR-10}
	\end{minipage}
	\begin{minipage}[t]{3.4in}
		\centering
		\includegraphics[trim=0mm 0mm 0mm 0mm, clip,width=3.4in]{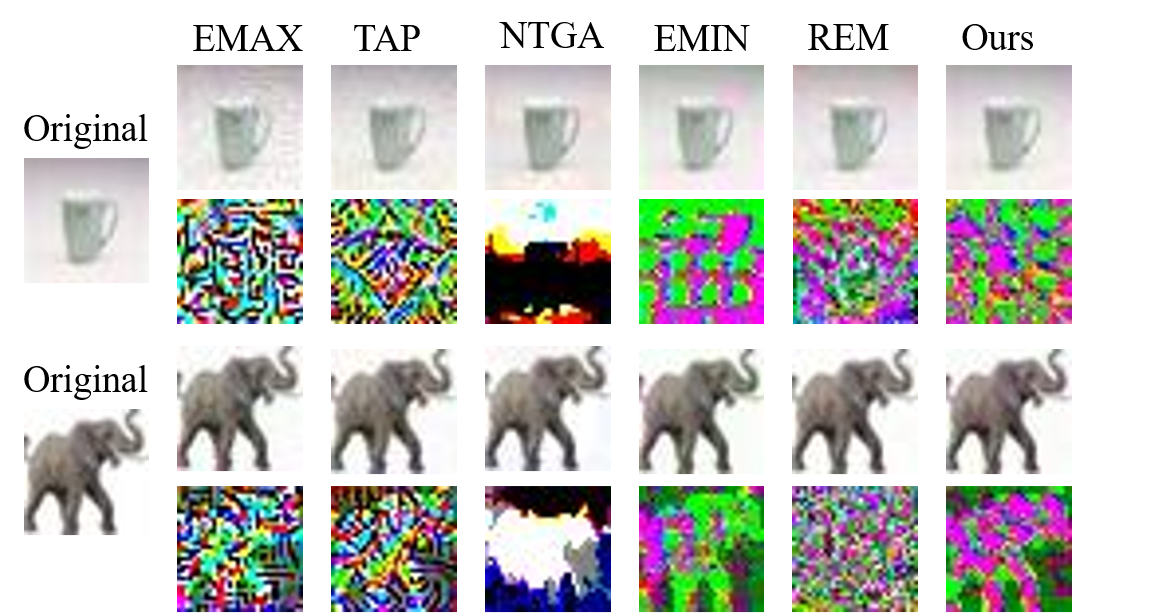}
		\centerline{\footnotesize (b) CIFAR-100}
	\end{minipage}\\
	\begin{minipage}[t]{3.5in}
		\centering
		\includegraphics[trim=0mm 0mm 0mm 0mm, clip,width=3.5in]{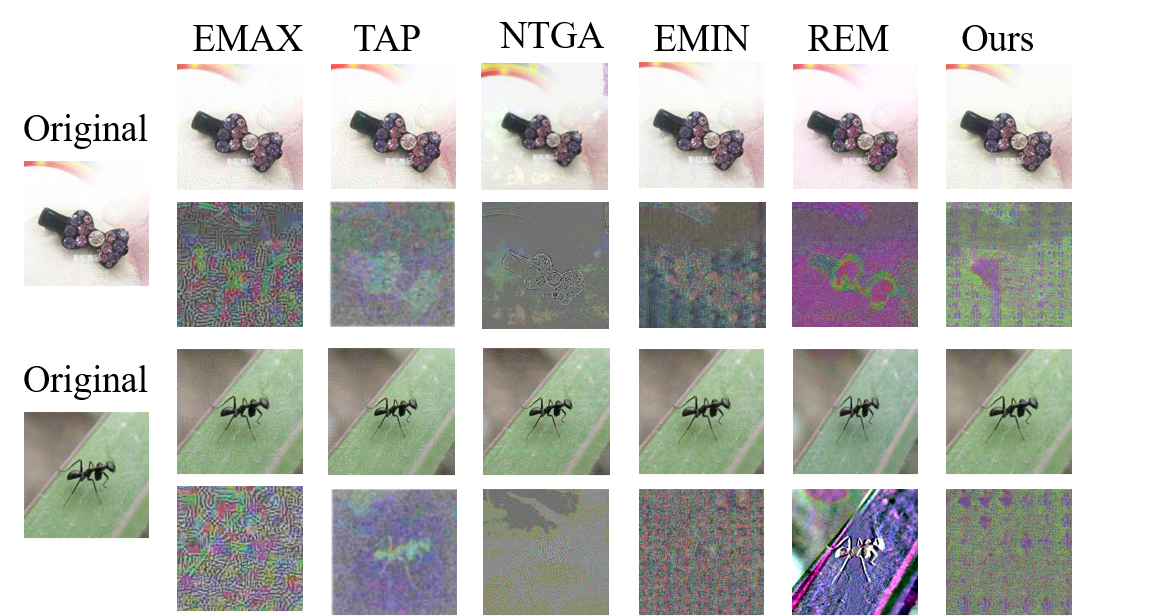}\\
		\centerline{\footnotesize (c) Mini-ImageNet}
	\end{minipage}
	\begin{minipage}[t]{3.4in}
		\centering
		\includegraphics[trim=0mm 0mm 0mm 0mm, clip,width=3.4in]{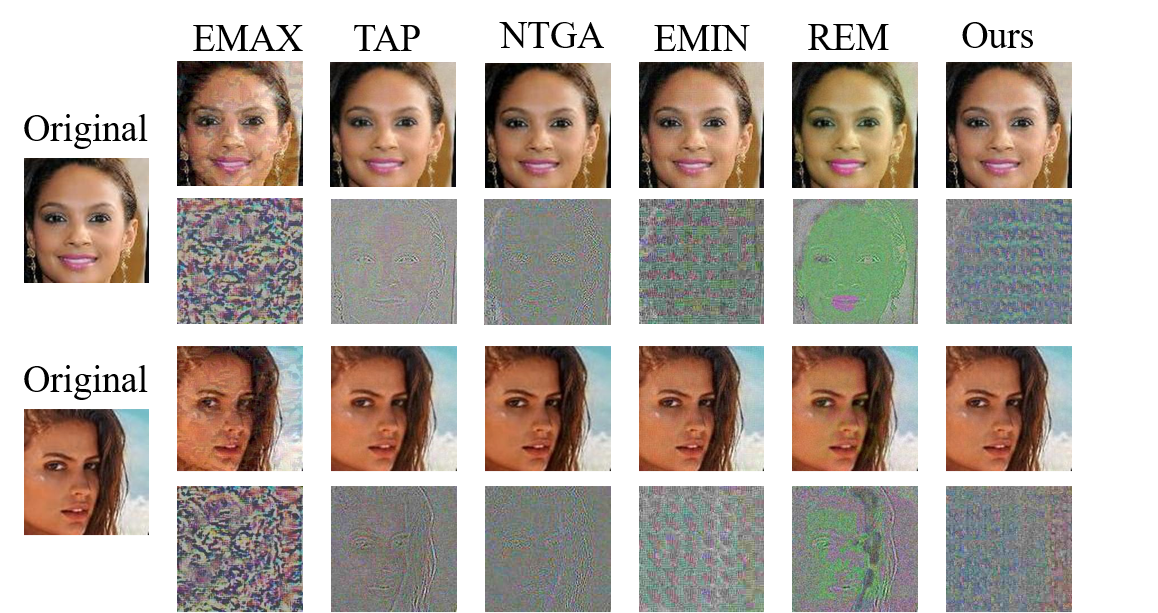}\\
		\centerline{\footnotesize (d) VGG-Face}
	\end{minipage}
	\caption{Visualization results of different types of defensive noise and corresponding unlearnable samples of EMAX \cite{madry2017towards}, TAP \cite{fowl2021adversarial}, NTGA \cite{yuan2021neural}, EMIN \cite{huang2021unlearnable}, REM \cite{fu2022robust}, and \sys. For each dataset, we randomly select two samples as the example.} \label{fig:poisoned}
\end{figure*}

\begin{table*}[tt]
	\caption{Test accuracy of models trained on data protected by \sys and baseline defensive noises in the non-data-augmentation scenario. \sys and baselines use sample-wise noise.}
	\label{tab:non-com}
	\centering
	\footnotesize
	\begin{tabular}{c|cccccccccccccccc}
		\toprule
		\multirow{2}{*}{\shortstack{Model}} &\multicolumn{9}{c}{CIFAR-10}  \\
        
		&\multicolumn{1}{c}{\color{black}Unprotected} & \multicolumn{1}{c}{\color{black}GaussNoise} &\multicolumn{1}{c}{\color{black}EMAX \cite{madry2017towards}} 
            &\multicolumn{1}{c}{\color{black}UTAP \cite{fowl2021adversarial}}
		&\multicolumn{1}{c}{\color{black}CTAP \cite{fowl2021adversarial}} &\multicolumn{1}{c}{\color{black}NTGA \cite{yuan2021neural}} & \multicolumn{1}{c}{\color{black}EMIN \cite{huang2021unlearnable}}& \multicolumn{1}{c}{\color{black}REM \cite{fu2022robust}}& \multicolumn{1}{c}{\color{black}\sys}\\
		\midrule
        VGG-16&92.66\%&92.49\%&90.91\%&85.90\%&34.70\%&	42.13\%&25.59\%&29.61\%& \textbf{14.66\%}\\
        ResNet-18&94.09\%&94.11\%&91.10\%&84.72\%&20.74\%&33.88\%&	25.08\%&24.15\%& \textbf{12.85\%}\\
        ResNet-50&94.38\%&93.56\%&91.78\%&85.35\%&18.82\%&20.55\%&19.19\%&20.50\%& \textbf{11.74\%}\\
        DenseNet-121&94.89\%&94.76\%&87.89\%&79.25\%&18.43\%&31.09\%&21.70\%&23.55\%& \textbf{11.09}\%\\
        WRN$\_$34$\_$10&95.52\%&95.61\%&89.84\%&82.76\%&18.77\%&25.52\%&21.38\%&24.50\%& \textbf{11.94\%}\\
        \hline
        \multirow{2}{*}{\shortstack{Model}} &\multicolumn{9}{c}{CIFAR-100}  \\
        
		&\multicolumn{1}{c}{\color{black}Unprotected} & \multicolumn{1}{c}{\color{black}GaussNoise} &\multicolumn{1}{c}{\color{black}EMAX \cite{madry2017towards}} 
           &\multicolumn{1}{c}{\color{black}UTAP \cite{fowl2021adversarial}}
		&\multicolumn{1}{c}{\color{black}CTAP \cite{fowl2021adversarial}}  &\multicolumn{1}{c}{\color{black}NTGA \cite{yuan2021neural}} & \multicolumn{1}{c}{\color{black}EMIN \cite{huang2021unlearnable}}& \multicolumn{1}{c}{\color{black}REM \cite{fu2022robust}}& \multicolumn{1}{c}{\color{black}\sys}\\
		\midrule
        VGG-16&69.43\%&69.30\%&66.16\%&63.26\%&62.45\%&35.86\%&10.81\%&15.10\%&	 \textbf{9.35\%}\\
        ResNet-18&71.84\%&71.22\%&68.10\%&66.15\%&64.00\%&22.41\%&15.19\%&13.18\%& \textbf{12.62\%}\\
        ResNet-50&71.69\%&69.61\%&69.79\%&66.10\%&65.31\%&19.74\%& 13.24\%&\textbf{11.91\%}&12.63\%\\
        DenseNet-121&73.04\%&72.67\%&71.50\%&67.32\%&67.66\%&28.58\%&13.64\%&13.37\%& \textbf{13.23\%}\\
        WRN$\_$34$\_$10&75.77\%&75.05\%&73.45\%&70.67\%&68.50\%&18.57\%&10.95\%&11.90\%& \textbf{7.46\%}\\
        \hline
        \multirow{2}{*}{\shortstack{Model}} &\multicolumn{9}{c}{Mini-ImageNet}  \\
        
		&\multicolumn{1}{c}{\color{black}Unprotected} & \multicolumn{1}{c}{\color{black}GaussNoise} &\multicolumn{1}{c}{\color{black}EMAX \cite{madry2017towards}} 
           &\multicolumn{1}{c}{\color{black}UTAP \cite{fowl2021adversarial}}
		&\multicolumn{1}{c}{\color{black}CTAP \cite{fowl2021adversarial}}  &\multicolumn{1}{c}{\color{black}NTGA \cite{yuan2021neural}} & \multicolumn{1}{c}{\color{black}EMIN \cite{huang2021unlearnable}}& \multicolumn{1}{c}{\color{black}REM \cite{fu2022robust}}& \multicolumn{1}{c}{\color{black}\sys}\\
		\midrule
        ResNet-18&75.69\%&59.37\%&57.45\%&61.39\%&21.00\%&50.76\%&14.98\%&26.81\%& \textbf{3.72\%}\\
        ResNet-50&74.56\%&65.55\%&57.25\%&60.37\%&18.70\%&52.20\%&11.67\%&23.43\%& \textbf{4.48\%}\\
        DenseNet-121&68.34\%&59.60\%&59.43\%&54.74\%&20.12\%&57.94\%&16.81\%&24.02\%& \textbf{6.00\%}\\
        EfficientNet&76.07\%&68.45\%&58.54\%&52.28\%&30.33\%&53.22\%&15.45\%&33.97\%& \textbf{2.25\%}\\
        ResNext-50&71.11\%&60.69\%&57.75\%&45.14\%&17.62\%&54.36\%&22.96\%&33.40\%& \textbf{2.90\%}\\
        \hline
        \multirow{2}{*}{\shortstack{Model}} &\multicolumn{9}{c}{VGG-Face}  \\
        
		&\multicolumn{1}{c}{\color{black}Unprotected} & \multicolumn{1}{c}{\color{black}GaussNoise} &\multicolumn{1}{c}{\color{black}EMAX \cite{madry2017towards}} 
            &\multicolumn{1}{c}{\color{black}UTAP \cite{fowl2021adversarial}}
		&\multicolumn{1}{c}{\color{black}CTAP \cite{fowl2021adversarial}}  &\multicolumn{1}{c}{\color{black}NTGA \cite{yuan2021neural}} & \multicolumn{1}{c}{\color{black}EMIN \cite{huang2021unlearnable}}& \multicolumn{1}{c}{\color{black}REM \cite{fu2022robust}}& \multicolumn{1}{c}{\color{black}\sys}\\
		\midrule
        ResNet-18&93.59\%&94.24\%&69.88\%&84.73\%&65.76\%&87.54\%&1.46\%&3.21\%& \textbf{0.28\%}\\
        ResNet-50&93.09\%&94.28\%&60.56\%&85.61\%&70.73\%&90.48\%&1.49\%&4.32\%& \textbf{0.19\%}\\
        DenseNet-121&93.62\%&95.80\%&61.36\%&85.92\%&72.54\%&91.36\%&1.57\%&5.53\%& \textbf{0.25\%}\\
        EfficientNet&95.56\%&96.20\%&61.53\%&87.63\%&73.25\%&88.94\%&1.53\%&3.46\%& \textbf{0.27\%}\\
        ResNext-50&94.54\%&93.65\%&62.42\%&84.32\%&67.56\%&86.75\%&1.65\%&4.21\%& \textbf{0.17\%}\\
		\bottomrule
	\end{tabular}
\vspace{-0.4cm}
\end{table*}

\begin{table*}[tt]
\vspace{-0.4cm}
	\caption{Test accuracy of models trained on data protected by \sys and baseline defensive noises in the data-augmentation scenario. \sys and baselines use sample-wise noise. }
	\label{tab:com-with}
	\centering
	\footnotesize
	\begin{tabular}{c|cccccccccccccccc}
		\toprule
		\multirow{2}{*}{\shortstack{Model}} &\multicolumn{9}{c}{CIFAR-10}  \\
        
		&\multicolumn{1}{c}{\color{black}Unprotected} & \multicolumn{1}{c}{\color{black}Random} &\multicolumn{1}{c}{\color{black}EMAX \cite{madry2017towards}} 
           &\multicolumn{1}{c}{\color{black}UTAP \cite{fowl2021adversarial}}
		&\multicolumn{1}{c}{\color{black}CTAP \cite{fowl2021adversarial}} &\multicolumn{1}{c}{\color{black}NTGA \cite{yuan2021neural}} & \multicolumn{1}{c}{\color{black}EMIN \cite{huang2021unlearnable}}& \multicolumn{1}{c}{\color{black}REM \cite{fu2022robust}}& \multicolumn{1}{c}{\color{black}\sys}\\
		\midrule
        VGG-16&93.86\%&93.68\%&90.33\%&85.28\%&50.02\%&89.95\%&62.69\%&47.58\%& \textbf{29.59\%}\\
        ResNet-18&94.49\%&92.79\%&89.42\%&85.79\%&40.20\%&90.71\%&56.05\%&43.25\%& \textbf{28.84\%}\\
        ResNet-50&94.52\%&90.80\%&83.96\%&80.61\%&38.95\%&90.25\%&	51.01\%&41.24\%& \textbf{28.07\%}\\
        DenseNet-121&95.28\%&93.63\%&87.31\%&78.43\%&39.04\%&80.91\%&57.32\%&44.39\%& \textbf{31.32\%}\\
        WRN$\_$34$\_$10&96.54\%&96.17\%&90.43\%&84.00\%&42.30\%&82.55\%&66.14\%&50.38\%& \textbf{30.73\%}\\
        \hline
        \multirow{2}{*}{\shortstack{Model}} &\multicolumn{9}{c}{CIFAR-100}  \\
		&\multicolumn{1}{c}{\color{black}Unprotected} & \multicolumn{1}{c}{\color{black}Random} &\multicolumn{1}{c}{\color{black}EMAX \cite{madry2017towards}} 
            &\multicolumn{1}{c}{\color{black}UTAP \cite{fowl2021adversarial}}
		&\multicolumn{1}{c}{\color{black}CTAP \cite{fowl2021adversarial}}  &\multicolumn{1}{c}{\color{black}NTGA \cite{yuan2021neural}} & \multicolumn{1}{c}{\color{black}EMIN \cite{huang2021unlearnable}}& \multicolumn{1}{c}{\color{black}REM \cite{fu2022robust}}& \multicolumn{1}{c}{\color{black}\sys}\\
		\midrule
        VGG-16&65.06\%&64.04\%&60.31\%&58.26\%&54.68\%&63.15\%&40.36\%&30.70\%& \textbf{22.60\%}\\
        ResNet-18&75.01\%&72.15\%&69.59\%&67.44\%&64.91\%&52.30\%&39.92\%&27.07\%& \textbf{25.10\%} \\
        ResNet-50&73.42\%&72.62\%&69.17\%&68.51\%&63.38\%&62.64\%&42.45\%&25.74\%& \textbf{24.04\%}\\
        DenseNet-121&76.06\%&73.76\%&70.35\%&67.00\%&64.26\%&62.63\%&40.32\%&32.74\%& \textbf{26.48\%}\\
        WRN$\_$34$\_$10&79.25\%&76.92\%&73.88\%&71.11\%&71.12\%&54.41\%&46.03\%&30.25\%& \textbf{26.40\%}\\
        \hline
        \multirow{2}{*}{\shortstack{Model}} &\multicolumn{9}{c}{Mini-ImageNet}  \\
        
	&\multicolumn{1}{c}{\color{black}Unprotected} & \multicolumn{1}{c}{\color{black}Random} &\multicolumn{1}{c}{\color{black}EMAX \cite{madry2017towards}} 
           &\multicolumn{1}{c}{\color{black}UTAP \cite{fowl2021adversarial}}
		&\multicolumn{1}{c}{\color{black}CTAP \cite{fowl2021adversarial}}  &\multicolumn{1}{c}{\color{black}NTGA \cite{yuan2021neural}} & \multicolumn{1}{c}{\color{black}EMIN \cite{huang2021unlearnable}}& \multicolumn{1}{c}{\color{black}REM \cite{fu2022robust}}& \multicolumn{1}{c}{\color{black}\sys}\\
		\midrule
        ResNet-18&77.66\%&67.90\%&59.99\%&63.26\%&37.69\%&64.28\%&58.06\%&44.46\%& \textbf{11.69\%}\\
        ResNet-50&77.87\%&70.84\%&58.29\%&60.92\%&36.42\%&60.56\%&63.51\%&43.60\%& \textbf{14.27\%}\\
        DenseNet-121&70.50\%&69.35\%&54.59\%&54.02\%&35.12\%&63.06\%&58.84\%&42.52\%& \textbf{11.96\%}\\
        EfficientNet&72.53\%&68.64\%&62.19\%&51.79\%&40.74\%&66.42\%&53.08\%&53.95\%& \textbf{13.67\%}\\
        ResNext-50&74.30\%&68.56\%&57.50\%&45.74\%&34.44\%&58.23\%&60.78\%&47.92\%& \textbf{12.80\%}\\
        \hline
        \multirow{2}{*}{\shortstack{Model}} &\multicolumn{9}{c}{VGG-Face}  \\
        
		&\multicolumn{1}{c}{\color{black}Unprotected} & \multicolumn{1}{c}{\color{black}Random} &\multicolumn{1}{c}{\color{black}EMAX \cite{madry2017towards}} 
            &\multicolumn{1}{c}{\color{black}UTAP \cite{fowl2021adversarial}}
		&\multicolumn{1}{c}{\color{black}CTAP \cite{fowl2021adversarial}}  &\multicolumn{1}{c}{\color{black}NTGA \cite{yuan2021neural}} & \multicolumn{1}{c}{\color{black}EMIN \cite{huang2021unlearnable}}& \multicolumn{1}{c}{\color{black}REM \cite{fu2022robust}}& \multicolumn{1}{c}{\color{black}\sys}\\
		\midrule
        ResNet-18&97.18\%&97.21\%&89.65\%&97.02\%&80.66\%&94.35\%&42.63\%&36.86\%& \textbf{23.10\%}\\
        ResNet-50&97.21\%&97.61\%&89.66\%&97.00\%&82.57\%&96.57\%&48.75\%&46.43\%& \textbf{28.36\%}\\
        DenseNet-121&93.02\%&95.78\%&90.73\%&96.49\%&85.10\%&93.78\%&41.14\%&40.78\%& \textbf{26.11\%}\\
        EfficientNet&97.29\%&90.91\%&86.34\%&96.42\%&83.11\%&95.32\%&40.84\%&42.76\%& \textbf{20.79\%}\\
        ResNext-50&97.73\%&95.12\%&87.83\%&97.67\%&82.10\%&92.90\%&35.75\%&37.43\%& \textbf{18.97\%}\\
		\bottomrule
	\end{tabular}
\vspace{-0.4cm}
\end{table*}


\begin{table}[tt]
	\caption{Test accuracy of models trained on data protected by \sys and baseline defensive noises in the non-data-augmentation scenario. \sys and baselines use class-wise noise.}
	\label{tab:com-class-no}
	\centering
	\footnotesize 
	\setlength\tabcolsep{2pt}
 	\begin{tabular}{ll|cccc}
		\toprule
        
		 \multirow{1}{*}{\shortstack{Dataset}} &\multirow{1}{*}{\shortstack{Model}}
           &\multicolumn{1}{c}{\color{black}Unprotected}& \multicolumn{1}{c}{\color{black}EMIN }&\multicolumn{1}{c}{\color{black}EMAX }& \multicolumn{1}{c}{\color{black}\sys}\\
		\midrule
        \multirow{5}{*}{\shortstack{\color{black}CIFAR-10}}&VGG-16&92.66\%&19.75\%&27.89\%& \textbf{11.53\%}\\
        &ResNet-18&94.09\%&12.21\%&10.85\%& \textbf{10.69\%}\\
        &ResNet-50&94.38\%&12.49\%&12.74\%& \textbf{11.78\%}\\
        &DenseNet-121&94.89\%&11.76\%&13.27\%& \textbf{9.40\%}\\
        &WRN$\_$34$\_$10&95.52\%&15.64\%&15.44\%& \textbf{10.46\%}\\
        \midrule
        \multirow{5}{*}{\shortstack{\color{black}CIFAR-100}}&VGG-16&69.43\%&2.25\%&	37.21\%& \textbf{1.58\%}\\
        &ResNet-18&71.84\%&5.17\%&29.74\%& \textbf{1.69\%}\\
        &ResNet-50&71.69\%&1.97\%&24.44\%& \textbf{1.55\%}\\
        &DenseNet-121&73.04\%&1.61\%&23.93\%& \textbf{1.12\%}\\
        &WRN$\_$34$\_$10&75.77\%&1.89\%&23.31\%& \textbf{1.71\%}\\
        \midrule
        \multirow{5}{*}{\shortstack{\color{black}Mini-ImageNet}}
        &ResNet-18&75.69\%&1.40\%&7.81\%& \textbf{1.34\%}\\
        &ResNet-50&74.56\%&1.95\%&7.92\%& \textbf{1.52\%}\\
        &DenseNet-121&68.34\%&2.05\%&6.38\%& \textbf{1.59\%}\\
        &MobileNet&76.34\%&1.29\%&4.05\%& \textbf{1.46\%}\\
        &ResNext-50&71.11\%&2.11\%&6.14\%& \textbf{2.04\%}\\
        \midrule
        \multirow{5}{*}{\shortstack{\color{black}VGG-Face}}
        &ResNet-18&93.59\%&1.28\%&1.28\%& \textbf{0.51\%}\\
        &ResNet-50&93.09\%&0.39\%&1.28\%& \textbf{0.32\%}\\
        &DenseNet-121&93.62\%&0.67\%&0.48\%& \textbf{0.26\%}\\
        &ResNext-50&88.72\%&0.70\%&1.28\%& \textbf{0.28\%}\\
        &EfficientNet&95.56\%& \textbf{0.43\%}&0.51\%&0.48\%\\
		\bottomrule
	\end{tabular}
\end{table}
\begin{table}[tt]
	\caption{Test accuracy of models trained on data protected by \sys and baseline defensive noises in the data-augmentation scenario. \sys and baselines use class-wise noise.}
	\label{tab:com-class-with}
	\centering
	\footnotesize
	\setlength\tabcolsep{2pt}
	\begin{tabular}{ll|cccc}
		\toprule
        
		 \multirow{1}{*}{\shortstack{Dataset}} &\multirow{1}{*}{\shortstack{Model}}
           &\multicolumn{1}{c}{\color{black}Unprotected}& \multicolumn{1}{c}{\color{black}EMIN }&\multicolumn{1}{c}{\color{black}EMAX }& \multicolumn{1}{c}{\color{black}\sys}\\
		\midrule
        \multirow{5}{*}{\shortstack{\color{black}CIFAR-10}}&VGG-16&93.86\%&54.41\%&65.80\%& \textbf{27.23\%}\\
        &ResNet-18&94.49\%&41.45\%&51.71\%& \textbf{25.48\%}\\
        &ResNet-50&94.52\%&51.74\%&47.30\%& \textbf{23.68\%}\\
        &DenseNet-121&95.38\%&45.79\%&51.41\%& \textbf{26.58\%}\\
        &WRN$\_$34$\_$10&96.54\%&37.14\%&55.06\%& \textbf{25.45\%}\\
       \midrule
        \multirow{5}{*}{\shortstack{\color{black}CIFAR-100}}&VGG-16&65.06\%&14.98\%&62.26\%& \textbf{13.50\%}\\
        &ResNet-18&75.01\%&24.53\%&59.64\%& \textbf{22.48\%}\\
        &ResNet-50&73.42\%&19.17\%&49.71\%& \textbf{13.17\%}\\
        &DenseNet-121&76.06\%&14.06\%&51.02\%& \textbf{12.09\%}\\
        &WRN$\_$34$\_$10&79.25\%&16.36\%&48.88\%& \textbf{13.59\%}\\
       \midrule
        \multirow{5}{*}{\shortstack{\color{black}Mini-ImageNet}}
        &ResNet-18&77.66\%&22.68\%&44.26\%& \textbf{7.67}\%\\
        &ResNet-50&77.87\%&22.48\%&46.92\%& \textbf{8.19\%}\\
        &DenseNet-121&70.50\%&24.33\%&38.35\%& \textbf{6.42\%}\\
        &EfficientNet&72.53\%&18.56\%&42.22\%& \textbf{4.06\%}\\
        &ResNext-50&74.30\%&29.97\%&44.59\%& \textbf{8.43\%}\\
       \midrule
        \multirow{5}{*}{\shortstack{VGG-Face}}&ResNet-18&97.18\%&19.87\%&21.25\%& \textbf{4.13\%}\\
        &ResNet-50&97.21\%&26.22\%&33.22\%& \textbf{5.56\%}\\
        &DenseNet-121&96.02\%&17.15\%&22.98\%& \textbf{9.36\%}\\
        &ResNext-50&93.77\%&28.54\%&33.15\%& \textbf{1.28\%}\\
        &EfficientNet&97.29\%&10.86\%&18.28\%& \textbf{2.72\%}\\
		\bottomrule
	\end{tabular}
\end{table}

\begin{table}[tt]
	\caption{Ablation study in the non-data-augmentation scenario. }
	\label{tab:ablation1}
	\centering
	\footnotesize
	\setlength\tabcolsep{0.5pt}
	\begin{tabular}{ll|cccc}
		\toprule
        
		  \multirow{2}{*}{\shortstack{Dataset}}&\multirow{2}{*}{\shortstack{Model}}
           & \multirow{2}{*}{\shortstack{Unprotected}}& \multirow{2}{*}{\shortstack{Base\\Noise}}& \multirow{2}{*}{\shortstack{Base+\\Non-local}}&\multirow{2}{*}{\shortstack{ Base+Non-local+\\Adaptive$\_$Stepsize}}\\
           &&&&&\\
		\midrule
        \multirow{5}{*}{\shortstack{\color{black}CIFAR-10}}&VGG-16&92.66\%&24.54\%&	16.75\%& \textbf{14.66\%}\\
        &ResNet-18&94.09\%&18.98\%&15.43\%& \textbf{12.85\%}\\
        &ResNet-50&94.38\%&14.54\%&14.34\%& \textbf{11.74\%}\\
        &DenseNet-121&94.89\%&18.14\%&16.96\%& \textbf{11.09\%}\\
        &WRN$\_$34$\_$10&95.52\%&17.55\%&16.90\%& \textbf{11.94\%}\\
       \midrule
        \multirow{5}{*}{\shortstack{\color{black}CIFAR-100}}&VGG-16&69.43\%&13.30\%&12.03\%& \textbf{9.35\%}\\
        &ResNet-18&71.84\%&20.60\%&15.25\%& \textbf{12.62\%}\\
        &ResNet-50&71.69\%&16.94\%&11.96\%& \textbf{11.63\%}\\
        &DenseNet-121&73.04\%&23.43\%&17.79\%& \textbf{13.23\%}\\
        &WRN$\_$34$\_$10&75.77\%&20.11\%&16.20\%& \textbf{7.46\%}\\
        \midrule
         
        \multirow{5}{*}{\shortstack{\color{black}Mini-ImageNet}}
        &ResNet-18&75.69\%&9.55\%&6.03\%& \textbf{3.72\%}\\
        &ResNet-50&74.56\%&8.78\%&4.99\%& \textbf{4.48\%}\\
        &DenseNet-121&68.34\%&10.86\%&6.00\%& \textbf{4.90\%}\\
        &EfficientNet&76.07\%&12.97\%&6.46\%& \textbf{2.25\%}\\
        &ResNext-50&71.11\%&8.91\%&4.49\%& \textbf{2.90\%}\\
        \midrule
        \multirow{5}{*}{\shortstack{\color{black}VGG-Face}}
        &ResNet-18&93.59\%&0.32\%&0.29\%& \textbf{0.28\%}\\
        &ResNet-50&93.09\%&0.54\%&0.36\%& \textbf{0.19\%}\\
        &DenseNet-121&93.62\%&0.60\%&0.39\%& \textbf{0.25\%}\\
        &ResNext-50&88.72\%&0.38\%&0.33\%& \textbf{0.27\%}\\
        &EfficientNet&95.56\%&0.45\%&0.19\%& \textbf{0.17\%}\\
		\bottomrule
	\end{tabular}
\end{table}

\begin{table}[tt]
\vspace{-0.4cm}
	\caption{Ablation study in the data-augmentation scenario. }
	\label{tab:ablation2}
	\centering
	\scriptsize
	\setlength\tabcolsep{1pt}
	\begin{tabular}{lc|cccc}
		\toprule
        
		  \multirow{2}{*}{\shortstack{Dataset}}&\multirow{2}{*}{\shortstack{Model}}
           & \multirow{2}{*}{\shortstack{Unprotected}}& \multirow{2}{*}{\shortstack{Base\\Noise}}& \multirow{2}{*}{\shortstack{Base+\\Non-local}}&\multirow{2}{*}{\shortstack{ Base+Non-local+\\Adaptive$\_$Stepsize}}\\
           &&&&&\\
		\midrule
        \multirow{5}{*}{\shortstack{\color{black}CIFAR-10}}&VGG-16&93.86\%&46.76\%&40.89\%& \textbf{29.59\%}\\
        &ResNet-18&94.49\%&42.24\%&41.92\%& \textbf{28.84\%}\\
        &ResNet-50&94.52\%&44.66\%&40.62\%& \textbf{28.07\%}\\
        &DenseNet-121&95.28\%&45.27\%&43.56\%& \textbf{31.32\%}\\
        &WRN$\_$34$\_$10&96.54\%&44.07\%&43.66\%& \textbf{30.73\%}\\
       \midrule
        \multirow{5}{*}{\shortstack{\color{black}CIFAR-100}}&VGG-16&65.06\%&27.72\%&27.31\%& \textbf{22.60\%}\\
        &ResNet-18&75.01\%&32.25\%&29.11\%& \textbf{25.10\%}\\
        &ResNet-50&73.42\%&33.21\%&24.43\%& \textbf{24.04\%}\\
        &DenseNet-121&76.06\%&30.29\%&29.67\%& \textbf{26.48\%}\\
        &WRN$\_$34$\_$10&79.25\%&34.15\%&28.98\%& \textbf{26.40\%}\\
        \midrule
         
        \multirow{5}{*}{\shortstack{\color{black}Mini-ImageNet}}
        &ResNet-18&77.66\%&25.34\%&18.64\%& \textbf{11.69\%}\\
        &ResNet-50&77.87\%&26.90\%&25.17\%& \textbf{14.27\%}\\
        &DenseNet-121&70.50\%&24.90\%&19.39\%& \textbf{11.96\%}\\
        &EfficientNet&72.53\%&24.67\%&24.10\%& \textbf{13.67\%}\\
        &ResNext-50&74.30\%&21.88\%&14.39\%& \textbf{12.80\%}\\
        \midrule
        \multirow{5}{*}{\shortstack{\color{black}VGG-Face}}
        &ResNet-18&97.18\%&27.68\%&26.43\%& \textbf{23.10\%}\\
        &ResNet-50&97.21\%&33.69\%&33.54\%& \textbf{31.36\%}\\
        &DenseNet-121&96.02\%&29.13\%&26.89\%& \textbf{26.11\%}\\
        &ResNext-50&93.77\%&30.45\%&26.65\%& \textbf{20.79\%}\\
        &EfficientNet&97.29\%&21.76\%&20.99\%& \textbf{8.97\%}\\
		\bottomrule
	\end{tabular}
\end{table}

\begin{table}[tt]
	\caption{Effectivenss of \sys and baselines under different data augmentation strategies. }
	\label{tab:dataaug}
	\centering
	\scriptsize
	\setlength\tabcolsep{1.5pt}
	\begin{tabular}{l|cccccccccccccccc}
		\toprule
		\multirow{1}{*}{\shortstack{Dataset}}&\multirow{1}{*}{\shortstack{Method}}
            &\multicolumn{1}{c}{\color{black}Unprotected}
		&\multicolumn{1}{c}{\color{black}EMIN}
           &\multicolumn{1}{c}{\color{black}CTAP}
		&\multicolumn{1}{c}{\color{black}NTGA} &\multicolumn{1}{c}{\color{black}REM}& \multicolumn{1}{c}{\color{black}\sys}\\
		\midrule
        \multirow{5}{*}{\shortstack{\color{black}CIFAR-10}}&FastAA&94.81\%&44.00\%&42.71\%&86.96\%&38.40\%& \textbf{22.91\%}\\
        &Mixup&95.03\%&33.65\%&32.78\%&44.66\%&31.58\%& \textbf{21.15\%}\\
        &PuzzleMix&95.37\%&40.52\%&45.53\%&34.85\%&40.43\%& \textbf{17.93\%}\\
        &FD $^1$&92.91\%&31.17\%&68.11\%&81.58\%&32.67\%& \textbf{16.13\%}\\
        &DeepAA&94.49\%&56.05\%&40.20\%&90.71\%&43.25\%& \textbf{28.84\%}\\
        \midrule
        \multirow{1}{*}{\shortstack{Dataset}}&\multirow{1}{*}{\shortstack{Method}}
            &\multicolumn{1}{c}{\color{black}Unprotected}
		&\multicolumn{1}{c}{\color{black}EMIN}
           &\multicolumn{1}{c}{\color{black}CTAP}
		&\multicolumn{1}{c}{\color{black}NTGA} &\multicolumn{1}{c}{\color{black}REM}& \multicolumn{1}{c}{\color{black}\sys}\\
		\midrule
        \multirow{5}{*}{\shortstack{\color{black}CIFAR-100}}&FastAA&73.64\%&32.27\%&64.92\%&51.47\%&24.72\%& \textbf{17.12\%}\\
        &Mixup&77.19\%&18.02\%&64.69\%&28.74\%&17.61\%& \textbf{14.73\%
        }\\
        &PuzzleMix&68.65\%&28.57\%&62.73\%&24.37\%&21.12\%& \textbf{12.55\%}\\
        &FD$^1$&68.48\%&22.08\%&60.39\%&54.69\%&25.05\%& \textbf{21.57\%}\\
        &DeepAA&75.01\%&39.92\%&64.91\%&52.30\%&27.07\%& \textbf{22.60\%}\\
        \midrule
        \multirow{1}{*}{\shortstack{Dataset}}&\multirow{1}{*}{\shortstack{Method}}
            &\multicolumn{1}{c}{\color{black}Unprotected}
		&\multicolumn{1}{c}{\color{black}EMIN}
           &\multicolumn{1}{c}{\color{black}CTAP}
		&\multicolumn{1}{c}{\color{black}NTGA} &\multicolumn{1}{c}{\color{black}REM}& \multicolumn{1}{c}{\color{black}\sys}\\
		\midrule
        \multirow{5}{*}{\shortstack{\color{black}Mini-ImageNet}}
        &FastAA&76.60\%&18.94\%&50.32\%&58.49\%&38.73\%& \textbf{13.77\%}\\
        &Mixup&76.89\%&17.79\%&44.97\%&50.77\%&36.98\%& \textbf{5.42\%}\\
        &PuzzleMix&76.46\%&18.14\%&60.64\%&55.78\%&64.95\%& \textbf{5.02\%}\\
        &FD$^1$&77.11\%&21.40\%&46.12\%&58.80\%&36.13\%& \textbf{11.10\%}\\
        &DeepAA&77.66\%&58.06\%&63.26\%&64.28\%&44.46\%& \textbf{11.69\%}\\
        \midrule
        \multirow{1}{*}{\shortstack{Dataset}}&\multirow{1}{*}{\shortstack{Method}}
            &\multicolumn{1}{c}{\color{black}Unprotected}
		&\multicolumn{1}{c}{\color{black}EMIN}
           &\multicolumn{1}{c}{\color{black}CTAP}
		&\multicolumn{1}{c}{\color{black}NTGA} &\multicolumn{1}{c}{\color{black}REM}& \multicolumn{1}{c}{\color{black}\sys}\\
		\midrule
        \multirow{5}{*}{\shortstack{\color{black}VGG-Face}}&FastAA&94.97\%&20.08\%&72.78\%&93.59\%&15.74\%& \textbf{3.30\%}\\
        &Mixup&94.49\%&7.96\%&65.91\%&90.12\%&6.76\%& \textbf{0.57\%}\\
        &PuzzleMix&94.36\%&8.78\%&66.36\%&91.62\%&5.87\%& \textbf{0.65\%}\\
        &FD$^1$&95.40\%&8.65\%&67.84\%&92.92\%&7.34\%& \textbf{0.65\%}\\
        &DeepAA&97.18\%&42.63\%&80.66\%&94.35\%&36.86\%& \textbf{23.10\%}\\
		\bottomrule
	\end{tabular}
  \begin{tablenotes}
\item  {\footnotesize $^1$ Short for Feature distillation.}

 \end{tablenotes}
\vspace{-0.4cm}
\end{table}

\begin{table}[tt]

	\caption{Impact of surrogate model structure on protection ability in the non-data-augmentation scenario.}
	\label{tab:sourcemodel1}
	\centering
	\scriptsize
	\setlength\tabcolsep{1pt}
	\begin{tabular}{lc|cccccccccccccc}
		\toprule
        
		\multirow{1}{*}{\shortstack{Dataset}}&\multirow{1}{*}{\shortstack{Unprotected}}
         &\multicolumn{1}{c}{\color{black}VGG-16} & \multicolumn{1}{c}{\color{black}ResNet-18} &\multicolumn{1}{c}{\color{black}ResNet-50} 
           &\multicolumn{1}{c}{\color{black}DenseNet-121}
		\\
		\midrule
        \multirow{1}{*}{\shortstack{\color{black}CIFAR-10}}&94.09\%&22.79\%&12.85\%&22.61\%&26.82\%\\
	  \multirow{1}{*}{\shortstack{\color{black}CIFAR-100}}&71.84\%&11.11\%&12.62\%&13.65\%&16.39\%\\

        \midrule
         \multirow{1}{*}{\shortstack{Dataset}}&\multirow{1}{*}{\shortstack{Unprotected}}
         & \multicolumn{1}{c}{\color{black}ResNet-18} &\multicolumn{1}{c}{\color{black}ResNet-50} 
           &\multicolumn{1}{c}{\color{black}DenseNet-121}&\multicolumn{1}{c}{\color{black}ResNext-50} \\
           \midrule
           \multirow{1}{*}{\shortstack{\color{black}Mini-ImageNet}}&75.69\%&3.72\%&9.06\%&10.95\%&8.95\%\\
	     \midrule
           \multirow{1}{*}{\shortstack{Dataset}}&\multirow{1}{*}{\shortstack{Unprotected}} & \multicolumn{1}{c}{\color{black}ResNet-18} &\multicolumn{1}{c}{\color{black}ResNet-50} 
           &\multicolumn{1}{c}{\color{black}DenseNet-121}&\multicolumn{1}{c}{\color{black}EffientNet} \\
           \midrule
	    \multirow{1}{*}{\shortstack{\color{black}VGG-Face}}&93.59\%&0.28\%&0.54\%&0.46\%&0.43\%\\
		\bottomrule
	\end{tabular}
	
\vspace{-0.4cm}
\end{table}

\begin{table}[tt]
\vspace{-0.4cm}
	\caption{Impact of surrogate model structure on protection ability in the data-augmentation scenario.}
	\label{tab:sourcemodel2}
	\centering
	\footnotesize
	\setlength\tabcolsep{0.5pt}
	\begin{tabular}{lc|cccccccccccccc}
		\toprule
        
		\multirow{1}{*}{\shortstack{Dataset}}&\multirow{1}{*}{\shortstack{Unprotected}}
         &\multicolumn{1}{c}{\color{black}VGG-16} & \multicolumn{1}{c}{\color{black}ResNet-18} &\multicolumn{1}{c}{\color{black}ResNet-50} 
           &\multicolumn{1}{c}{\color{black}DenseNet-121}
		\\
		\midrule
        \multirow{1}{*}{\shortstack{\color{black}CIFAR-10}}
       &94.49\%&29.82\%&30.84\%&29.49\%&30.41\%\\

        \multirow{1}{*}{\shortstack{\color{black}CIFAR-100}}
        &75.01\%&24.26\%&25.10\%&27.50\%&29.68\%\\

        \midrule
         \multirow{1}{*}{\shortstack{Dataset}}&\multirow{1}{*}{\shortstack{Unprotected}}
         & \multicolumn{1}{c}{\color{black}ResNet-18} &\multicolumn{1}{c}{\color{black}ResNet-50} 
           &\multicolumn{1}{c}{\color{black}DenseNet-121}&\multicolumn{1}{c}{\color{black}ResNext-50} \\
           \midrule
           \multirow{1}{*}{\shortstack{\color{black}Mini-ImageNet}}&77.66\%&11.69\%&26.12\%&25.78\%&18.76\%\\
           \midrule
           \multirow{1}{*}{\shortstack{Dataset}}&\multirow{1}{*}{\shortstack{Unprotected}} & \multicolumn{1}{c}{\color{black}ResNet-18} &\multicolumn{1}{c}{\color{black}ResNet-50} 
           &\multicolumn{1}{c}{\color{black}DenseNet-121}&\multicolumn{1}{c}{\color{black}EffientNet} \\
           \midrule
	    \multirow{1}{*}{\shortstack{\color{black}VGG-Face}}&97.18\%&23.10\%&28.65\%&29.03\%&23.33\%\\
		\bottomrule
	\end{tabular}
\end{table}

\begin{table}[tt]
	\caption{Impact of $\epsilon$ on the effectiveness of \sys in the non-data-augmentation scenario.}
	\label{tab:noise1}
	\centering
	\footnotesize
	\setlength\tabcolsep{5pt}
	\begin{tabular}{cc|ccccccccccccccc}
		\toprule
        
		Dataset&\multicolumn{1}{c}{\color{black}Unprotected} & \multicolumn{1}{c}{\color{black}$\epsilon=4$} & \multicolumn{1}{c}{\color{black}$\epsilon=8$}& \multicolumn{1}{c}{\color{black}$\epsilon=12$}& \multicolumn{1}{c}{\color{black}$\epsilon=16$}
		\\
		\midrule
        \multirow{1}{*}{\shortstack{\color{black}CIFAR-10}}&94.09\%&14.03\%&12.85\%&10.41\%&9.87\%\\
        \multirow{1}{*}{\shortstack{\color{black}CIFAR-100}}&71.84\%&42.17\%&12.62\%&8.95\%&8.05\%\\

        \multirow{1}{*}{\shortstack{\color{black}Mini-ImageNet}}&75.69\%&9.27\%&3.72\%&3.50\%&3.02\%\\

        \multirow{1}{*}{\shortstack{\color{black}VGG-Face}}&93.59\%&2.45\%&0.65\%&0.46\%&0.28\%\\
		\bottomrule
	\end{tabular}
\end{table}

\begin{table}[tt]
	\caption{Impact of $\epsilon$ on the effectiveness of \sys in the data-augmentation scenario.}
	\label{tab:noise2}
	\centering
	\footnotesize
	\setlength\tabcolsep{5pt}
	\begin{tabular}{cc|ccccccccccccccc}
		\toprule
        
		Dataset&\multicolumn{1}{c}{\color{black}Unprotected} & \multicolumn{1}{c}{\color{black}$\epsilon=4$} & \multicolumn{1}{c}{\color{black}$\epsilon=8$}& \multicolumn{1}{c}{\color{black}$\epsilon=12$}& \multicolumn{1}{c}{\color{black}$\epsilon=16$}
		\\
		\midrule
        \multirow{1}{*}{\shortstack{\color{black}CIFAR-10}}
        &94.49\%&33.56\%&30.84\%&30.50\%&29.05\%\\
	
        \multirow{1}{*}{\shortstack{\color{black}CIFAR-100}}&75.01\%&60.07\%&25.10\%&23.68\%&19.60\%\\

        \multirow{1}{*}{\shortstack{\color{black}Mini-ImageNet}}
        &77.66\%&28.50\%&11.69\%&10.05\%&9.04\%\\

        \multirow{1}{*}{\shortstack{\color{black}VGG-Face}}&97.18\%&95.60\%&70.68\%&55.56\%&23.10\%\\
		\bottomrule
	\end{tabular}
\vspace{-0.4cm}
\end{table}

\begin{table*}[tt]
\vspace{-0.4cm}
	\caption{Impact of the number of protection labels on the protection ability in non-data-augmentation scenarios. }
	\label{tab:number1}
	\centering
	\footnotesize
	\begin{tabular}{lll|ccc|ccccc}
		\toprule
        
		  \multirow{2}{*}{\shortstack{Dataset}}&\multirow{2}{*}{\shortstack{Protected class}} &\multirow{2}{*}{\shortstack{Test accuracy of \#}}&\multicolumn{3}{c|}{Sample-wise}&\multicolumn{3}{c}{Class-wise}\\
           &&& \multicolumn{1}{c}{\color{black}EMAX}&\multicolumn{1}{c}{\color{black}EMIN}& \multicolumn{1}{c|}{\color{black}\sys}& \multicolumn{1}{c}{\color{black}EMAX}&\multicolumn{1}{c}{\color{black}EMIN}& \multicolumn{1}{c}{\color{black}\sys}\\
		\midrule
        \multirow{6}{*}{\shortstack{\color{black}CIFAR-10}}
         &\multirow{2}{*}{\shortstack{\color{black}Class 0 $\sim$ 1}}&Unprotected label&94.50\%&94.41\%&94.24\%&94.49\%&94.36\%&94.71\%\\
        &&Protected label&10.85\%&7.65\%& \textbf{5.35\%}&3.40\%&11.30\%& \textbf{0.60\%} \\
        \cline{2-9}
        & \multirow{2}{*}{\shortstack{\color{black}Class 0 $\sim$ 4}}&Unprotected label&97.72\%&97.22\%&97.48\%&97.42\%&97.28\%&	97.80\%\\
        &&Protected label&2.75\%&	2.96\%& \textbf{1.24\%}&0.36\%&1.32\%& \textbf{0.08\%}\\
        \cline{2-9}
        & \multirow{2}{*}{\shortstack{\color{black}Class 0 $\sim$ 7}}&Unprotected label&98.60\%&	98.20\%&98.25\%&98.50\%&98.60\%&	98.70\%\\
        &&Protected label&1.84\%&1.48\%& \textbf{0.44\%}&	0.89\%&0.68\%& \textbf{0.05\%}\\
        \hline
         \multirow{8}{*}{\shortstack{\color{black}CIFAR-100}}
         &\multirow{2}{*}{\shortstack{\color{black}Class 0 $\sim$ 9}}&Unprotected label&72.12\%&71.53\%&72.93\%&72.12\%&72.30\%&			71.71\%\\
        &&Protected label&18.36\%&22.90\%& \textbf{20.50\%}&29.40\%&10.80\%&			 \textbf{9.40\%}\\
        \cline{2-9}
        & \multirow{2}{*}{\shortstack{\color{black}Class 0 $\sim$ 19}}&Unprotected label&74.78\%&71.60\%&72.79\%&74.48\%&74.06\%&73.98\%\\
        &&Protected label&14.10\%&15.65\%&	 \textbf{12.10\%}&11.80\%&4.60\%& \textbf{3.85\%}\\
        \cline{2-9}
        & \multirow{2}{*}{\shortstack{\color{black}Class 0 $\sim$ 49}}&Unprotected label&78.02\%&74.24\%&77.54\%&78.06\%&78.40\%&77.54\%\\
        &&Protected label&11.90\%&9.48\%& \textbf{7.42\%}&12.26\%&5.24\%& \textbf{3.02\%}\\
        \cline{2-9}
        & \multirow{2}{*}{\shortstack{\color{black}Class 0 $\sim$ 79}}&Unprotected label&86.46\%&83.35\%&86.60\%&89.05\%&89.45\%&87.95\%\\
        &&Protected label&5.88\%&6.54\%& \textbf{5.52\%}&8.32\%&4.37\%& \textbf{2.32\%}\\
        \hline
         \multirow{6}{*}{\shortstack{\color{black}Mini-ImageNet}}
         &\multirow{2}{*}{\shortstack{\color{black}Class 0 $\sim$ 19}}&Unprotected label&73.56\%&72.54\%&73.47\%&73.40\%&73.79\%&73.94\% \\
        &&Protected label&23.25\%&28.05\%& \textbf{6.95\%}&27.60\%& \textbf{19.15\%}&19.60\% \\
        \cline{2-9}
        & \multirow{2}{*}{\shortstack{\color{black}Class 0 $\sim$ 49}}&Unprotected label&76.22\%&76.44\%&74.66\%&76.76\%&75.04\%&73.80\% \\
        &&Protected label&21.40\%&11.54\%& \textbf{3.64\%}&19.22\%&0.48\%& \textbf{0.20\%} \\
        \cline{2-9}
        & \multirow{2}{*}{\shortstack{\color{black}Class 0 $\sim$ 79}}&Unprotected label&86.15\%&86.45\%&83.40\%&86.00\%&85.10\%&84.00\% \\
        &&Protected label&2.59\%&4.61\%& \textbf{0.59\%}&7.84\%&0.10\%& \textbf{0.08\%}\\
        \hline
         \multirow{6}{*}{\shortstack{\color{black}VGG-Face}}
         &\multirow{2}{*}{\shortstack{\color{black}Class 0 $\sim$ 39}}&Unprotected label&94.61\%&94.61\%&94.44\%&94.51\%&94.39\%&94.54\%\\
        &&Protected label&74.55\%&9.77\%& \textbf{8.42\%}&0.29\%& \textbf{0.00\%}& \textbf{0.00\%}\\
        \cline{2-9}
        & \multirow{2}{*}{\shortstack{\color{black}Class 0 $\sim$ 99}}&Unprotected label&95.07\%&94.82\%&94.97\%&94.92\%&94.75\%&94.75\%\\
        &&Protected label&71.99\%&1.11\%& \textbf{0.98\%}&0.10\%&0.08\%& \textbf{0.00\%}\\
        \cline{2-9}
        & \multirow{2}{*}{\shortstack{\color{black}Class 0 $\sim$ 159}}&Unprotected label&95.98\%&95.29\%&94.43\%&94.15\%&94.22\%&	94.60\%\\
        &&Protected label&63.68\%&0.01\%& \textbf{0.00\%}&0.06\%&0.04\%& \textbf{0.00\%}\\
		\bottomrule
	\end{tabular}
\vspace{-0.4cm}
\end{table*}

\begin{table*}[tt]
	\caption{Impact of the number of protection labels on the protection ability in the data-augmentation scenario. }
	\label{tab:number2}
	\centering
	\footnotesize
	\begin{tabular}{lll|ccc|ccccc}
		\toprule
        
		  \multirow{2}{*}{\shortstack{Dataset}}&\multirow{2}{*}{\shortstack{Protected class}} &\multirow{2}{*}{\shortstack{Test accuracy of \#}}&\multicolumn{3}{c|}{Sample-wise}&\multicolumn{3}{c}{Class-wise}\\
           &&& \multicolumn{1}{c}{\color{black}EMAX}&\multicolumn{1}{c}{\color{black}EMIN}& \multicolumn{1}{c|}{\color{black}\sys}& \multicolumn{1}{c}{\color{black}EMAX}&\multicolumn{1}{c}{\color{black}EMIN}& \multicolumn{1}{c}{\color{black}\sys}\\
		\midrule
        \multirow{6}{*}{\shortstack{\color{black}CIFAR-10}}
         &\multirow{2}{*}{\shortstack{\color{black}Class 0 $\sim$ 1}}&Unprotected label&92.46\%&93.08\%&93.05\%&93.36\%&92.89\%&93.15\% \\
        &&Protected label&36.12\%&50.80\%& \textbf{28.65\%}&16.54\%&28.50\%& \textbf{11.50\%}\\
        \cline{2-9}
        & \multirow{2}{*}{\shortstack{\color{black}Class 0 $\sim$ 4}}&Unprotected label&97.30\%&96.64\%&96.32\%&97.14\%&96.80\%&96.52\%\\
        &&Protected label&31.25\%&26.19\%& \textbf{7.40\%}&	9.55\%&7.40\%& \textbf{3.12\%}\\
        \cline{2-9}
        & \multirow{2}{*}{\shortstack{\color{black}Class 0 $\sim$ 7}}&Unprotected label&98.50\%&97.90\%&97.45\%&98.10\%&97.80\%&97.65\%\\
        &&Protected label&4.64\%&13.20\%& \textbf{5.75\%}&	6.08\%&5.04\%& \textbf{1.34\%}\\
        
        \hline
         \multirow{8}{*}{\shortstack{\color{black}CIFAR-100}}
         &\multirow{2}{*}{\shortstack{\color{black}Class 0 $\sim$ 9}}&Unprotected label&75.22\%&74.62\%&74.68\%&74.89\%&72.61\%&			75.74\%\\
        &&Protected label&21.60\%&32.00\%& \textbf{22.40\%}&38.80\%&14.50\%&			 \textbf{14.00\%}\\
        \cline{2-9}
        & \multirow{2}{*}{\shortstack{\color{black}Class 0 $\sim$ 19}}&Unprotected label&76.26\%&75.59\%&76.11\%&76.02\%&76.29\%&76.40\%\\
        &&Protected label&14.80\%&25.95\%& \textbf{18.45\%}&26.95\%&7.75\%& \textbf{7.25\%}\\
        \cline{2-9}
        & \multirow{2}{*}{\shortstack{\color{black}Class 0 $\sim$ 49}}
        &Unprotected label&80.14\%&77.50\%&78.06\%&80.10\%&	79.60\%&79.16\%\\
        &&Protected label&12.23\%&21.54\%& \textbf{9.32\%}&16.06\%&	8.22\%& \textbf{6.36\%}\\
        \cline{2-9}
        & \multirow{2}{*}{\shortstack{\color{black}Class 0 $\sim$ 79}}&Unprotected label&87.56\%&85.80\%&85.70\%&90.45\%&87.94\%&84.85\%\\
        &&Protected label&9.95\%&18.59\%& \textbf{8.04\%
        }&13.38\%&5.86\%& \textbf{3.71\%}\\
        \hline
         \multirow{6}{*}{\shortstack{\color{black}Mini-ImageNet}}
         &\multirow{2}{*}{\shortstack{\color{black}Class 0 $\sim$ 19}}&Unprotected label&77.81\%&77.61\%&77.34\%&77.89\%&77.34\%&77.09\%\\
        &&Protected label&30.14\%&33.15\%& \textbf{12.00\%}&30.90\%&33.60\%& \textbf{11.20\%}\\
        \cline{2-9}
        & \multirow{2}{*}{\shortstack{\color{black}Class 0 $\sim$ 49}}&Unprotected label&80.20\%&79.44\%&79.82\%&80.14\%&80.14\%&78.80\%\\
        &&Protected label&27.10\%&25.38\%& \textbf{5.70\%}&25.28\%&16.20\%& \textbf{5.76\%}\\
        \cline{2-9}
        & \multirow{2}{*}{\shortstack{\color{black}Class 0 $\sim$ 79}}&Unprotected label&89.70\%&88.35\%&87.45\%&88.65\%&86.75\%&87.20\%\\
        &&Protected label&20.00\%&25.64\%& \textbf{2.50\%}&20.99\%&4.68\%& \textbf{1.98\%}\\
        \hline
         \multirow{6}{*}{\shortstack{\color{black}VGG-Face}}
         &\multirow{2}{*}{\shortstack{\color{black}Class 0 $\sim$ 39}}&Unprotected label&97.25\%&97.46\%&97.20\%&97.41\%&90.43\%&97.34\%\\
        &&Protected label&47.17\%&77.89\%& \textbf{36.38\%}&44.30\%&62.75\%& \textbf{24.19\%}\\
        \cline{2-9}
        & \multirow{2}{*}{\shortstack{\color{black}Class 0 $\sim$ 99}}&Unprotected label&97.63\%&97.76\%&97.81\%&97.83\%&97.77\%&97.57\% \\
        &&Protected label&29.13\%&72.55\%& \textbf{25.97\%}&22.84\%&51.20\%& \textbf{22.06\%}\\
        \cline{2-9}
        & \multirow{2}{*}{\shortstack{\color{black}Class 0 $\sim$ 159}}&Unprotected label&97.61\%&98.27\%&98.20\%&97.47\%&98.37\%&	97.47\% \\
        &&Protected label&20.83\%&48.94\%& \textbf{15.02\%}&10.77\%&39.31\%& \textbf{4.05\%}\\
		\bottomrule
	\end{tabular}
\end{table*}

\begin{table*}[tt]
\vspace{-0.4cm}
	\caption{Computing cost of \sys and baselines.}
	\label{tab:time}
	\centering
	\setlength\tabcolsep{8pt}
 \footnotesize
	\begin{tabular}{c|cccccccccccccccc}
		\toprule
		\multirow{1}{*}{\shortstack{Dataset}}& \multicolumn{1}{c}{\color{black}EMIN}& \multicolumn{1}{c}{\color{black}EMAX}
           &\multicolumn{1}{c}{\color{black}TAP} & \multicolumn{1}{c}{\color{black}REM}
		 &\multicolumn{1}{c}{\color{black}NTGA}& \multicolumn{1}{c}{\color{black}\sys}\\
		\midrule
        CIFAR-10&0.56h&0.30h+0.42h&	9.50h&38.00h+2.50h&17.67h
&0.50h+1.90h\\
        CIFAR-100&1.50h&0.50h+0.40h&10.70h&33.50h+3.50h&18.56h&0.65h+3.13h\\
        Mini-ImageNet&3.50h&2.24h+2.08h&16.00h&27.00h+22.00h&32.50h&1.10h+6.12h\\
        VGG-Face&5.60h&2.64h+2.88h&13.00h&18.75h+34.00h&38.35h&1.35h+10.25h\\
		\bottomrule
	\end{tabular}
\vspace{-0.4cm}
\end{table*}
\begin{table}[tt]
	\caption{Effectiveness of \sys and REM under adversarial training.}
	\label{tab:adversarialtraining}
	\centering
	\footnotesize
	\setlength\tabcolsep{3pt}
	\begin{tabular}{lcccccccccccccccc}
		\toprule
		\multirow{1}{*}{\shortstack{Dataset}}&\multirow{1}{*}{\shortstack{Method}}&$\rho$$^*$
            &\multicolumn{1}{c}{\color{black}Unprotected}
		& \multicolumn{1}{c}{\color{black}REM}
            & \multicolumn{1}{c}{\color{black}\sys} \\
	\midrule
  
        \multirow{3}{*}{\shortstack{\color{black}CIFAR-10}}&Super-convergence&-&88.20\%&43.24\%&14.16\%\\
        \cline{2-6}
        &\multirow{2}{*}{\shortstack{Propagate}}&1&93.74\%&10.10\%&12.90\%\\
        &&2&91.16\%&31.47\%&34.38\%\\
        \midrule
        
        \multirow{3}{*}{\shortstack{\color{black}CIFAR-100}}&
        Super-convergence&-&66.56\%&14.83\%&26.54\%\\
        \cline{2-6}
        &\multirow{2}{*}{\shortstack{Propagate}}&1&72.10\%&12.06\%&11.75\%\\
        &&2&68.89\%&14.69\%&28.18\%\\
        
	\midrule
  
        \multirow{3}{*}{\shortstack{\color{black}Mini-ImageNet}}&
       Super-convergence&-&69.18\%&6.78\%&9.32\%\\
       \cline{2-6}
        &Propagate&1&73.42\%&3.84\%&6.90\%\\
        &Propagate&2&70.26\%&4.44\%&8.17\%\\
        \midrule
       
        \multirow{3}{*}{\shortstack{\color{black}VGG-Face}}&        Super-convergence&-&94.66\%&70.77\%&32.35\%\\
        \cline{2-6}
        &\multirow{2}{*}{\shortstack{Propagate}}&1&91.68\%&17.72\%&35.26\%\\
        &&2&92.93\%&27.19\%&35.72\%\\
    
		\bottomrule
	\end{tabular}
  \begin{tablenotes}
\item  {\footnotesize $^*$ The hyperparameter $\rho$ refers to the perturbation radius of PGD attack.}

 \end{tablenotes}
\vspace{-0.5cm}
\end{table}

\subsection{Defensive Noise Generation}

The defensive noise is generated to make the attacker model have a small loss on the protected sample. Thus, the attacker model can hardly learn any constructive information from the image.
{We use cross-entropy as the loss function and the first-order optimization method PGD \cite{madry2017towards} to update the noise as}
\begin{equation}\label{23}
    \delta_{t+1} = \delta_{t}-\alpha*\mathrm{sign}(\nabla\mathcal{L}(f(\mathcal{A}(\mathbf{x}+\delta_t, \mathbf{p}),\mathbf{y}))),
\end{equation}
where $t$ is the current iteration, and $\alpha$ is the learning rate. 

The learning rate $\alpha$ determines the length of each step taken in the negative direction of the gradient during the gradient descent iteration process. If the learning rate is too small, the training process may be too slow to converge. If the learning rate is too large, the training loss may fluctuate and fail to decrease monotonically. Therefore, it is essential to dynamically adjust the learning rate according to the status of the current iteration. Traditionally, PGD uses a constant learning rate. Instead, we adjust the learning rate to be inversely proportional to the input gradient norm. Specifically, samples with larger gradient norms are given lower learning rates \cite{huang2022fast}, and vice versa. 
\begin{equation}
\begin{split}
   & n_{i, t}=\beta n_{i, t-1} + (1-\beta)||\nabla_{\mathbf{x}_{i, t-1}}  \mathcal{L} (x_{i, t-1}, y_i)||_2^2,\\
  & \alpha_t=\frac{\gamma}{c+\sqrt{\sum_i n_{i,t}}},
\end{split}
\end{equation}
where $x_{i,t}$ is $i$-th sample at the $t$-th iteration, $\beta$ is the momentum, $n_{i,t}$ is the moving average of the gradient norm, and $\gamma, c$ are hyperparameters to prevent $\alpha_t$ from being too large.

The defensive noise generation process is iterated until the training process of the surrogate model converges. Note that our noise generation algorithm can generate both sample-wise noise and class-wise noise. Sample-wise noise can be directly optimized via Equation~(\ref{23}). For class-wise noise, we compute and average the generated defensive noise of each sample in a given class. 

{ The defensive noise generation process is summarized in Algorithm~\ref{alg2}. }


\section{Evaluation Setup}
\subsection{Model and Datasets} 
We conduct experiments on four widely-used datasets, i.e., CIFAR-10 \cite{krizhevsky2009learning}, CIFAR-100 \cite{krizhevsky2009learning}, Mini-ImageNet \cite{imagentte}, and VGG-Face \cite{parkhi2015deep}. 
We utilize ResNet-18 to train the surrogate model on CIFAR-10, CIFAR-100, Mini-ImageNet, and VGG-Face by default. 

\textbf{CIFAR-10}. CIFAR-10 \cite{krizhevsky2009learning} consists of 60,000 color images, each of size 32 $\times$ 32 pixels, divided into 10 different classes. Each class contains 6,000 images. The images in CIFAR-10 are quite diverse and can be challenging to classify accurately due to variations in lighting, scale, orientation, and the presence of overlapping objects. We randomly select 50,000 samples from the dataset to form the training set, while the remaining 10,000 samples are reserved for the test set. We train the model on the training set for 100 epochs. The learning rate is 0.1, the batch size is 128, the weight\_decay is 0.0005, and the momentum stochastic gradient descent is 0.9.
A CosineAnnealingLR scheduler is used in the training process, with a $T\_max$ value of 100 and an $eta\_min$ value of 0.

\textbf{CIFAR-100}. 
CIFAR-100 \cite{krizhevsky2009learning} is an extension of the CIFAR-10 dataset. It consists of 60,000 color images, each of size 32 $\times$ 32 pixels, but unlike CIFAR-10, it contains 100 different classes. Each class in CIFAR-100 represents a fine-grained category, and there are 600 images per class. We divide the dataset into a training set and a test set, with 50,000 images for training and 10,000 images for testing. We train the model for 150 epochs. We set the learning rate as 0.1, the batch size as 128, and the momentum stochastic gradient descent as 0.9. 
A CosineAnnealingLR scheduler is used in the training process, with a $T\_max$ value of 150 and an $eta\_min$ value of 0.

\textbf{Mini-ImageNet.} Mini-ImageNet, a widely used subset of ImageNet \cite{imagentte}, holds significant popularity within the research community \cite{xiang2021patchguard,lu2021discriminator}. Mini-ImageNet is carefully selected to include diverse object categories and a range of image variations, making it suitable for evaluating and comparing different image classification models. Mini-ImageNet comprises a total of 60000 images, and we divide 40,000 images as the training set and 10,000 images as the test set. Each image has a high resolution with a dimension of 224 $\times$ 224. We train the model for 150 epochs. We set the learning rate as 0.1, the batch size as 64, the momentum of stochastic gradient descent as 0.9, and weight decay as 0.0005. 
A CosineAnnealingLR scheduler is used in the training process, with a $T\_max$ value of 150 and an $eta\_min$ value of 0.

\textbf{VGG-Face.} 
VGGFace \cite{parkhi2015deep} is a large-scale face recognition dataset that was created by researchers at the Visual Geometry Group (VGG). 
VGGFace contains a vast collection of face images, encompassing a wide range of identities from various sources. It includes approximately 2.6 million images of over 2,600 individuals, making it one of the largest face recognition datasets available. Each image has a high resolution with a dimension of 224 $\times$ 224.
In this paper, we randomly select 200 categories of images to form a dataset. There are 53,811 images for training and 10,762 images for testing. We resize the images to 3 $\times$ 224 $\times$ 224. We set the learning rate as 0.01. The mini-batch size is set as 64. 
A CosineAnnealingLR scheduler is used in the training process, with a $T\_max$ value of 150 and an $eta\_min$ value of 0.

\subsection{Evaluation Metrics}
In line with prior research \cite{fu2022robust, huang2021unlearnable}, we employ \emph{test accuracy} as a metric to evaluate the privacy-preserving capability of the noise. A lower test accuracy indicates that the model has acquired minimal knowledge from the training data, thereby suggesting a robust privacy protection ability of the noise.

In the experiments, we use sample-wise noise by default unless otherwise specified. We assume the commercial model trainer primarily uses DeepAA \cite{zheng2022deep}, a widely-used advanced automatic data augmentation strategy. The search space of the optimal augmentation operations is shown in Table~\ref{tab:auglist}. We also explore the effectiveness of \sys when the model trainer adopts other state-of-the-art data augmentation strategies, such as Mixup \cite{zhang2017mixup}, Feature distillation \cite{liu2019feature}, PuzzleMix \cite{kim2020puzzle}, and FastAA \cite{lim2019fast} in the experiments.

\subsection{Experiential Settings of Various Defensive Noise Generation Methods}
We compare \sys with several state-of-the-art data protection methods, including Gaussian noise, EMAX \cite{madry2017towards}, TAP \cite{fowl2021adversarial}, NTGA \cite{yuan2021neural}, EMIN \cite{huang2021unlearnable}, and REM \cite{fu2022robust}. 

\textbf{Gaussian noise.} 
We randomly generate noise independently for each training example or each class, sampling from the interval [$-\epsilon, \epsilon$].

\textbf{EMAX.} EMAX \cite{madry2017towards} is generated based on the gradients of a pre-trained model. We generate the noise for EMAX using a PGD-20 attack on a pre-trained ResNet-18 model applied to the training dataset, and the step size is set to 2.




\textbf{EMIN.} EMIN \cite{huang2021unlearnable} is generated according to Equation (\ref{Emin-equa}). In every epoch, we train the surrogate model for a few epochs to solve the outer minimization problem, and PGD is employed to solve the inner minimization problem. The number of training epochs is set to 10, and the PGD attack is parameterized with 10 rounds, each with a step size of 0.8. The stop condition error rate is $\lambda$ = 0.1 for sample-wise noise and $\lambda$ = 0.01 for class-wise noise.

\textbf{TAP.} TAP \cite{fowl2021adversarial} is generated by performing a targeted adversarial attack on the model trained with clean data. In this attack, we follow the hyperparameter settings as specified in the original work.

\textbf{NTGA.} NTGA \cite{yuan2021neural} is an efficient method enabling clean-label, black-box generalization attacks against Deep Neural Networks. This work is based on the development of Neural Tangent Kernels (NTKs). On each dataset, we randomly split 10\% of examples from the training dataset as a validation set to assist in the generation of noise.

\textbf{REM}. REM \cite{fu2022robust} follows Equation (\ref{REM-equa}) to generate the noise. We set the adversarial training perturbation radius as 2, as it is commonly used in the experiments of the original work. All the other hyperparameter settings follow the specifications of the original work.

\textbf{ARMOR}. 
In surrogate model construction, the non-local module is respectively added to the first two residual blocks of ResNet-18. For surrogate augmentation strategy selection, we choose ResNet-18 as our auxiliary model structure and pre-train the auxiliary model from scratch on the entire training dataset for 10 epochs. During each iteration, we train the auxiliary model with 10 batches of perturbed data, and then use it to update our surrogate augmentation strategy. In defensive noise generation, we follow the same process as EMIN. We train the surrogate model on 10 batches, and the PGD attack is parameterized with 10 rounds, each with a step size of 0.8. The stop condition is also the same as EMIN. 

Note that for baselines and \sys, the radius of defensive perturbation $\epsilon$ is set to 8/255. Based on prior studies in adversarial research, the defensive noise under this constraint is imperceptible to human observers.

All experiments are implemented in Python and run on a 14-core Intel(R) Xeon(R) Gold 5117 CPU @2.00GHz  and NVIDIA  GeForce RTX 3080 Ti GPU  machine running Ubuntu 18.04 system.

\section{Evaluation Results}

\subsection{Comparison with Baselines}
We compare \sys with several state-of-the-art data protection methods, including EMAX \cite{madry2017towards}, TAP \cite{fowl2021adversarial}, NTGA \cite{yuan2021neural}, EMIN \cite{huang2021unlearnable}, and REM \cite{fu2022robust}. 
We implement these baselines according to their open-source codes. We also use Gaussian noise as a baseline. 

We compare \sys with the baselines in both non-data-augmentation and data-augmentation scenarios. In the two scenarios, there are two kinds of noises, i.e., sample-wise noise and class-wise noise. Most of these methods generate sample-wise noise. Only EMAX \cite{madry2017towards} and EMIN \cite{huang2021unlearnable} also include class-wise noise. Thus, we only compare \sys with EMAX \cite{madry2017towards} and EMIN \cite{huang2021unlearnable} for the performance of class-wise noise. In the experiments, we use different model structures to test the effectiveness and transferability of the noise generated by \sys and baselines.

We first compare \sys with baselines in terms of sample-wise noise. The comparison results are shown in Table~\ref{tab:non-com} and Table~\ref{tab:com-with}. 
We can see that models trained on unlearnable data generated by \sys consistently yield the lowest test accuracy across various datasets and model architectures. 
Besides, random noise has almost no effect on protecting private data from being used to train a well-performed model.
In the non-data-augmentation scenario, take CIFAR-10 as an example, \sys brings the test accuracy from 92.66\% down to 14.66\% (VGG-16), from 94.09\% down to 12.85\% (ResNet-18), from 94.38\% down to 11.74\% (ResNet-50), from 94.89\% down to 11.09\% (DesNet-121), from 95.52\% down to 11.94\% (WRN$\_$34$\_$10), respectively. In contrast, the lowest model test accuracy of baseline methods is still as high as 25.59\% (VGG-16), 20.74\% (ResNet-18), 18.82\% (ResNet-50), 18.43\% (DenseNet-121), and 18.77\% (WRN$\_$34$\_$10). For the high-resolution datasets, \sys can significantly reduce the test accuracy of different models to less than 2.25\% (Mini-ImageNet) and less than 0.17\% (VGG-Face), which are more effective than the baselines. 
As shown in Table~\ref{tab:com-with}, we can see that the clean model accuracy improves after applying the data augmentation strategy, and most of the noises of baselines are ineffective in this case. For example, EMIN can reduce the test accuracy from 92.66\% down to 25.59\% without data augmentation for the CIFAR-10 dataset using VGG-16, however, it can only decrease the test accuracy from 93.86\% down to 62.69\% with data augmentation. In comparison, \sys can also successfully reduce the model test accuracy to less than 30\% in most cases under the data augmentation. These results show that \sys is robust to data augmentation.

We also compare \sys with baselines in terms of class-wise noise. The comparison results are shown in Table~\ref{tab:com-class-no} and Table~\ref{tab:com-class-with}. 
It is shown that \sys can also achieve the best protection ability in most cases for all datasets compared with state-of-the-art class-wise noise generation methods, especially for the data-augmentation scenario. We also discovered that class-wise noise is more effective than sample-wise noise in decreasing the model test accuracy. This discrepancy can be attributed to the explicit correlation exhibited by class-wise noise with the corresponding label. By learning this correlation, the model can effectively reduce training errors. Consequently, the model becomes unintentionally focused on learning the noise itself instead of capturing the true underlying content. However, class-wise noise is shown to be more likely to be exposed \cite{huang2021unlearnable}.
In the case of sample-wise noise, each individual sample is subjected to a distinct noise pattern, and there is no explicit correlation between the noise and the corresponding label. In this scenario, the model only tends to disregard low-error samples while attaching more importance to normal and high-error examples.

To explore the invisibility of the generated noise, we visualize the results of different types of defensive noise and the corresponding examples in Figure~\ref{fig:poisoned}. We can see the unlearnable samples generated by \sys retain a natural appearance, preserving the usability of the original image.

\subsection{Ablation Study}
In this section, we conduct an ablation study to examine the necessity of the base augmentation-resistant noise generation framework, the non-local module, and the dynamic step size adjustment algorithm. The results are shown in Table~\ref{tab:ablation1} and Table~\ref{tab:ablation2}. 
The column of ``Unprotected'' represents the model test accuracy on unprotected data. 
The column of ``Base Noise'' represents the test accuracy on protected data with only base noise of \sys without non-local module and dynamic step size adjustment. The ``Base+Non-local'' method uses the non-local module. The ``Base+Non-local+Adaptive$\_$Stepsize'' method is the complete data protection method of \sys. In this section, we use the sample-wise noise.

In both non-data-augmentation and data-augmentation scenarios, comparing ``Unprotected" and ``Base Noise", we can observe that our base noise generation framework can significantly decrease the model test accuracy. 
For example, when using the VGG-16 model, the clean test accuracy is 93.86\% for CIFAR-10 with data augmentation but reaches as low as 46.76\% using the ``Base Noise'' protection method. The reduction is more than 47.10\%. 

Compared with ``Base Noise'', the ``Base+Non-local'' method further decreases the model test accuracy, which shows the effectiveness of the non-local module. The success of the non-local module is that it encourages the surrogate model to also focus on information from other areas and avoid over-emphasizing localized features. 

It is shown that the dynamic step size adjustment algorithm can also further decrease the model test accuracy in almost all cases for all datasets. For example, when using the VGG-16 model, the ``Base+Non-local+Adaptive$\_$Stepsize'' protection method can further decrease test accuracy from 40.89\% to 29.59\% under the data augmentation scenario. The reduction is as high as 11.3\%.

\subsection{Performance under Different Data Augmentation Methods}  We explore the effectiveness of \sys and baselines when the model trainer adopts other state-of-the-art data augmentation strategies, such as Mixup \cite{zhang2017mixup}, Feature distillation \cite{liu2019feature}, PuzzleMix \cite{kim2020puzzle}, and FastAA \cite{lim2019fast}. Given that CTAP surpasses UTAP in data protection, we exclusively showcase the CTAP results here.
The results are shown in Table~\ref{tab:dataaug}. 

We can see that various data augmentation strategies can make the protected data learnable again, especially for TAP and NTGA. While EMIN and REM occasionally demonstrate resilience to data augmentation (i.e., reduce the test accuracy to less than 30\%), strategies like FastAA and PuzzMix can successfully improve the model test accuracy to larger than 40\% even 60\% for these methods. In comparison, \sys can successfully decrease the model test accuracy to less than 30\% (in most cases less than 20\%) for all the list advanced data augmentation methods for all datasets.

\subsection{Impact of Surrogate Model Structure}
By default, the surrogate model for the four datasets employs the ResNet-18 architecture. We explore whether \sys is also effective when using other surrogate model structures, such as VGG-16, ResNet-50, DenseNet-121, EfficientNet, and ResNext-50. Note that we set the commercial unauthorized model structure as ResNet-18.

The results are shown in Table~\ref{tab:sourcemodel1} and Table~\ref{tab:sourcemodel2}. It is shown that \sys is robust to the structures of the surrogate model. \sys can effectively reduce the test accuracy of the unauthorized commercial model regardless of the surrogate model structure.

\subsection{Impact of the $\epsilon$ Value}
In \sys, we use the first-order optimization method PGD \cite{madry2017towards} to update the noise $\delta$. In PGD, $\epsilon$ is the maximum perturbation allowed for the noise. We vary $\epsilon$ values and investigate its impact on \sys. The results are shown in Table~\ref{tab:noise1} and Table~\ref{tab:noise2}.

A higher $\epsilon$ offers enhanced protection but can make noise more noticeable. The choice of $\epsilon$ involves a trade-off between protection effectiveness and concealment. Extensive experimentation has revealed that an $\epsilon$ of 8 is ideal for CIFAR-10, CIFAR-100, and Mini-ImageNet datasets, making the noise nearly invisible. For the VGG-Face dataset, we found that a $\epsilon$ of 16 generates sufficiently natural images and can achieve the defense goal. The need for increased noise in the VGG-Face dataset arises from its characteristics. Images in VGG-Face are close together within the same category (inter-class distances) but far from different ones (inter-class distances). As a result, the VGG-Face dataset demonstrates a higher level of linear separability, leading to high classification accuracy \cite{ren2022transferable}. To effectively confuse the features between different categories, a higher $\epsilon$ is essential.

\subsection{Impact of Protection Label Number} 
We then explore the number of protection labels on the protection ability. In this case, defensive noise shields only a subset of labels, leaving samples from the remaining labels unprotected upon release. The test accuracy results of models trained on unprotected labels and the protected labels are shown in Table~\ref{tab:number1} and Table~\ref{tab:number2}. As for both \sys and baselines, we can see that as the number of protection labels rises, our defensive capability improves, leading to a decrease in the test accuracy of the protected label.

\subsection{Time Cost of Baselines and \sys}
We calculate the time costs for generating different types of defensive noise on different datasets. The results are shown in Table~\ref{tab:time}. 
The format of $t_1 + t_2$ represents that noise generation methods include two stages: model pre-training stage ($t_1$) and noise generation stage ($t_2$). We can see that \sys exhibits outstanding performance in terms of execution efficiency, surpassing several baselines such as TAP, NTGA, and REM by a significant margin. While \sys requires additional time compared to EMIN and EMAX, the noise generated by \sys exhibits significantly higher protection capabilities than that of EMIN and EMAX.

\subsection{Robustness to Adversarial Training}
Adversarial training is a technique proposed to enhance the robustness of deep learning models against adversarial examples. The idea behind adversarial training is to augment the training process by including adversarial examples during the training phase. The standard adversarial training \cite{madry2017towards} addresses a min-max problem, where the objective is to minimize the loss function while the adversarial examples are present. In our study, we investigate the resilience of \sys against adversarial training by considering commonly-used adversarial training strategies, namely Super-convergence \cite{smith2018superconvergence} and Propagate \cite{zhang2019propagate}. Super-convergence \cite{smith2018superconvergence} is based on the adversarial examples generated by FGSM \cite{kurakin2016adversarial}, and Propagate \cite{zhang2019propagate} is based on the adversarial examples generated by PGD \cite{madry2017towards}. We compared \sys with REM, whose defensive noise is specifically designed to resist adversarial training.
The results are shown in Table~\ref{tab:adversarialtraining}. 

Similarly to REM, we can see that \sys is also robust to adversarial training. \sys can effectively reduce the model test accuracy even if the model trainer adopts state-of-the-art adversarial training strategies. Take CIFAR as an example, after applying \cite{madry2017towards}, the test accuracy decreases significantly from 93.74\% to 12.90\%, highlighting the robustness of \sys against adversarial training.

\section{Conclusion}
This paper reveals the vulnerability of existing unlearnable examples to data augmentation, a widely-used pre-processing technique. To tackle this, we propose a novel unlearnable example generation framework, dubbed \sys, to safeguard data privacy against potential breaches arising from data augmentation. In \sys, we introduce a non-local module-assisted surrogate model to overcome the difficulty of having no access to the model training process. Additionally, we design a surrogate augmentation strategy selection algorithm that maximizes distribution alignment between augmented and non-augmented samples. Moreover, we propose using an adaptive learning rate adjustment algorithm to improve the defensive noise generation process.
Extensive experiments on various datasets, including CIFAR-10, CIFAR-100, Mini-ImageNet, and VGG-Face, verify the effectiveness and superiority of \sys in both sample-wise and class-wise noises. \sys has also demonstrated robustness to adversarial training.

\bibliographystyle{plain}
\bibliography{reference}




\section*{Biography Section}

\begin{IEEEbiography}[{\includegraphics[width=1in,height=1.25in,clip,keepaspectratio]{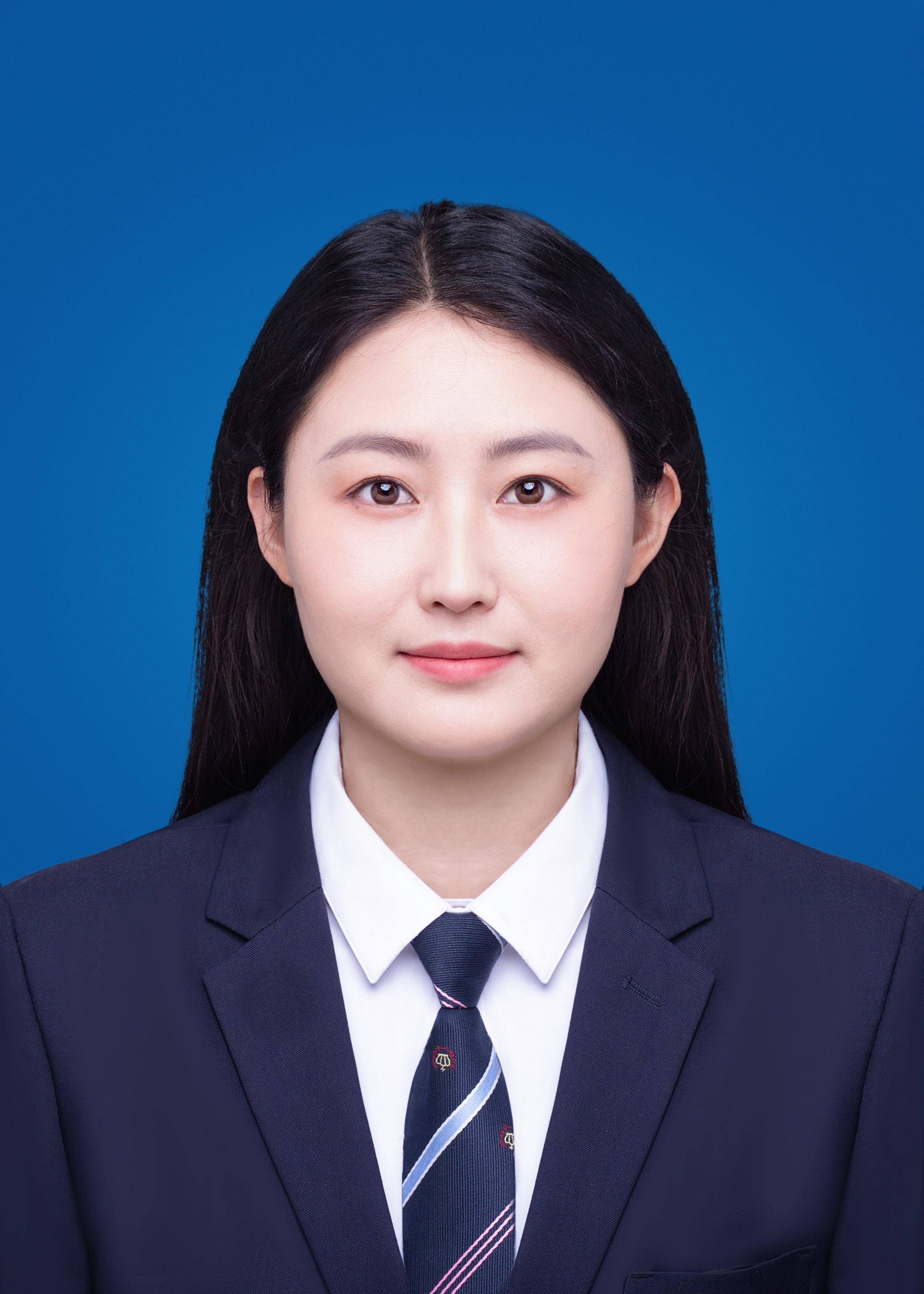}}]{Xueluan Gong} received her B.S. degree in Computer Science and Electronic Engineering from Hunan University in 2018. She received her Ph.D. degree in the School of Computer Science, Wuhan University, in 2023. Her research interests include network security and AI security. Her research has been published in multiple top-tier conferences and journals, such as IEEE S\&P, USENIX Security, NDSS, WWW, IJCAI, IEEE JSAC, IEEE TDSC, and IEEE TIFS. She also served as the reviewer of ICML, NeurIPS, ICLR, IEEE TIFS, IEEE TDSC, etc.
\end{IEEEbiography}

\begin{IEEEbiography}[{\includegraphics[width=1in,height=1.25in,clip,keepaspectratio]{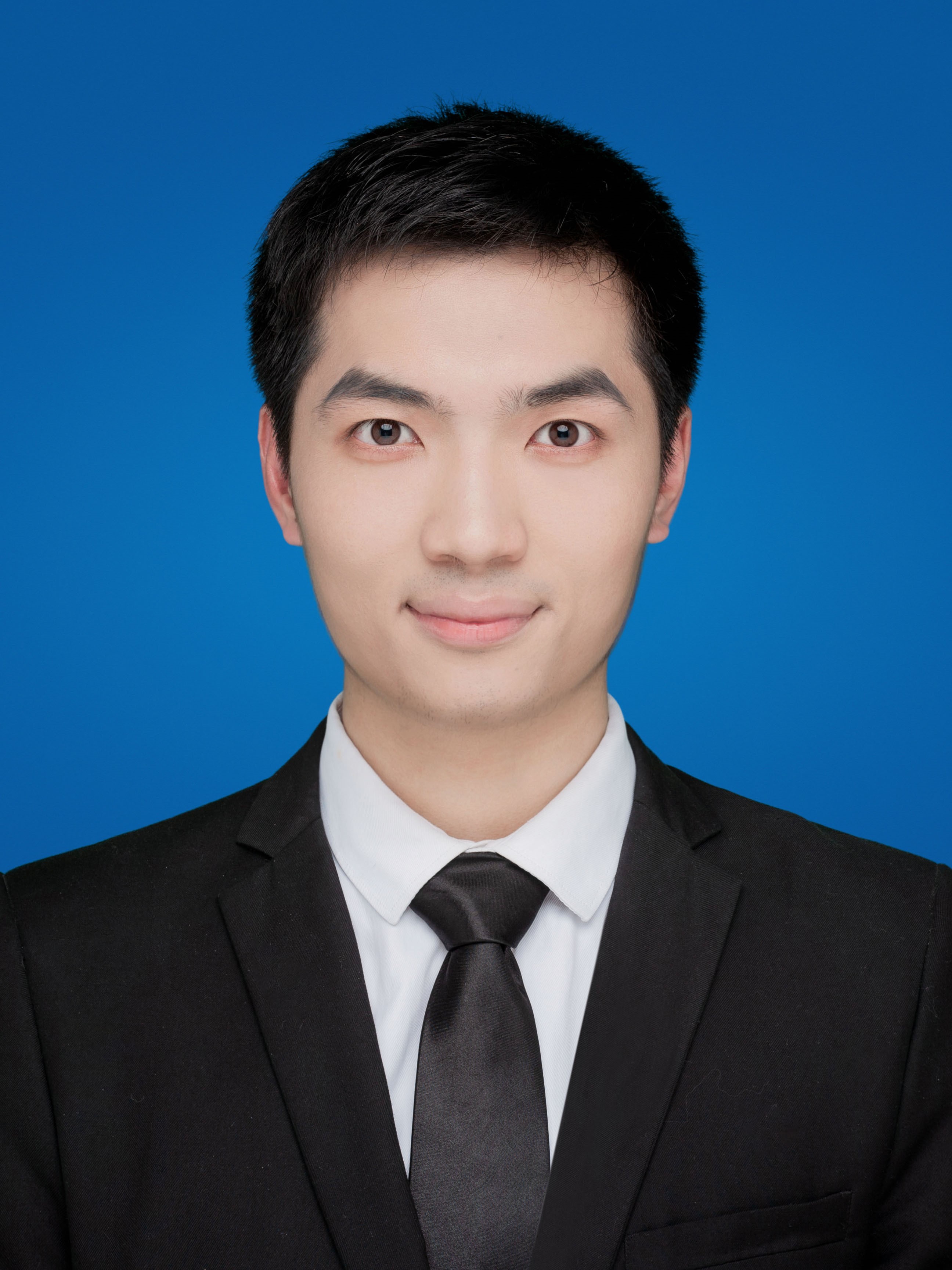}}]{Yuji Wang} is currently pursuing the B.E. at the School of Cyber Science and Engineering from Wuhan University, China. His research interests include AI security and information security.
\end{IEEEbiography}

\begin{IEEEbiography}[{\includegraphics[width=1in,height=1.25in,clip,keepaspectratio]{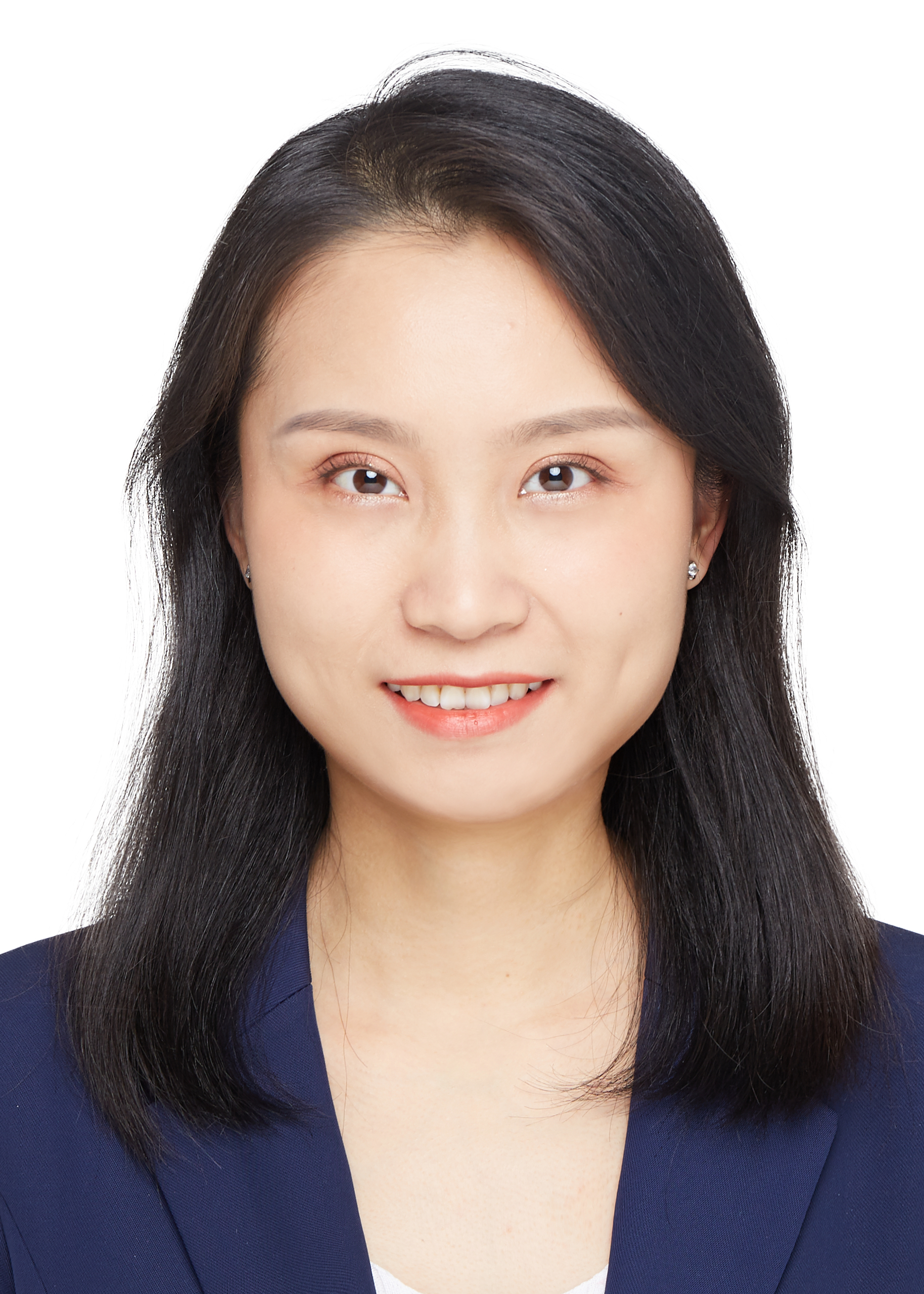}}]{Yanjiao Chen} received her B.E. degree in Electronic Engineering from Tsinghua University in 2010 and Ph.D. degree in Computer Science and Engineering from Hong Kong University of Science and Technology in 2015. 
She is currently a Bairen Researcher in Zhejiang University, China. Her research interests include spectrum management for Femtocell networks, network economics, network security, and Quality of Experience (QoE) of multimedia delivery/distribution.
\end{IEEEbiography}

\begin{IEEEbiography}[{\includegraphics[width=1in,height=1.2in,clip,keepaspectratio]{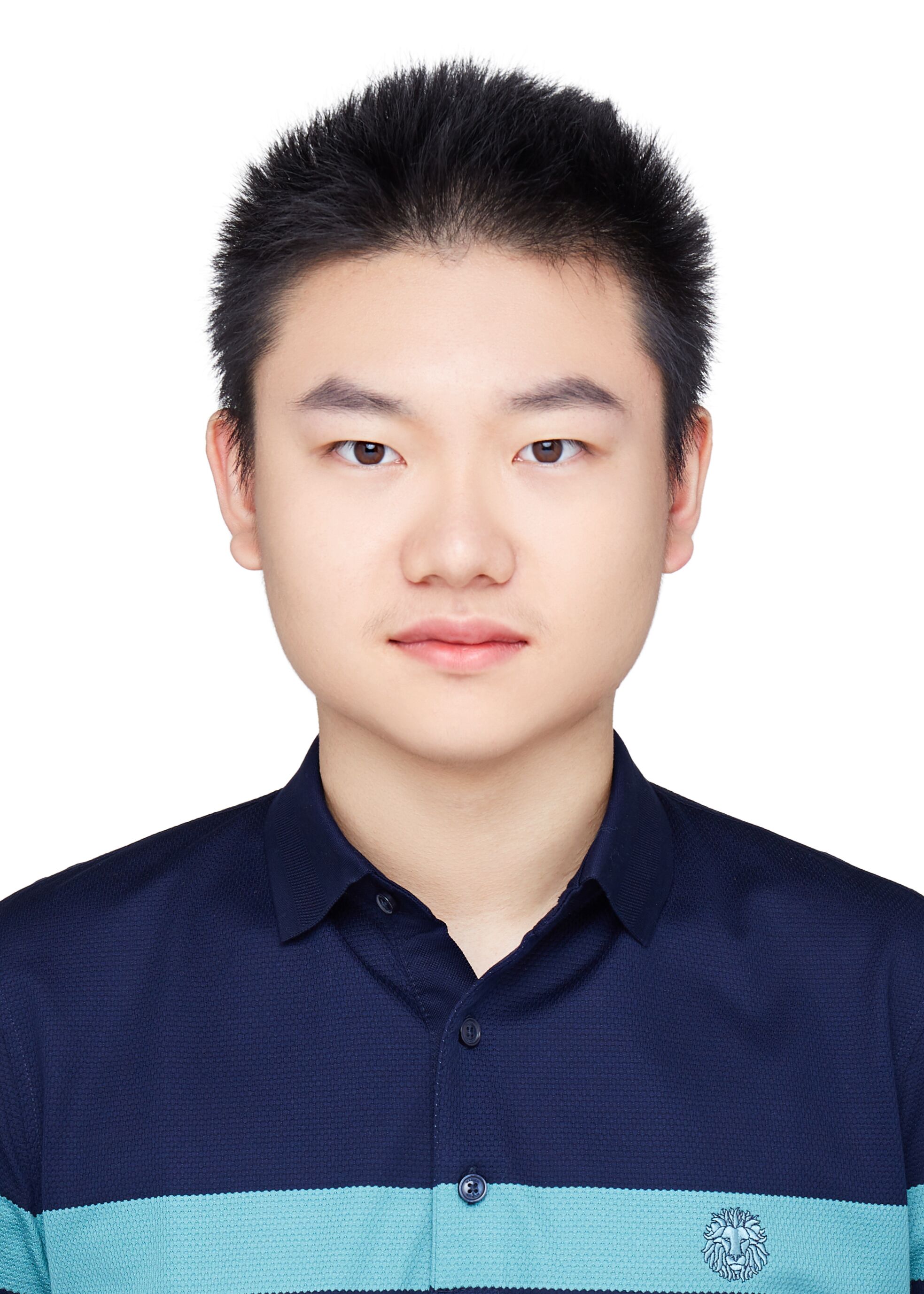}}]{Haocheng Dong} is currently pursuing the B.E. at the School of Cyber Science and Engineering from Wuhan University, China. His research interests include AI security and information security.
\end{IEEEbiography}

\begin{IEEEbiography}[{\includegraphics[width=1in,height=1.2in,clip,keepaspectratio]{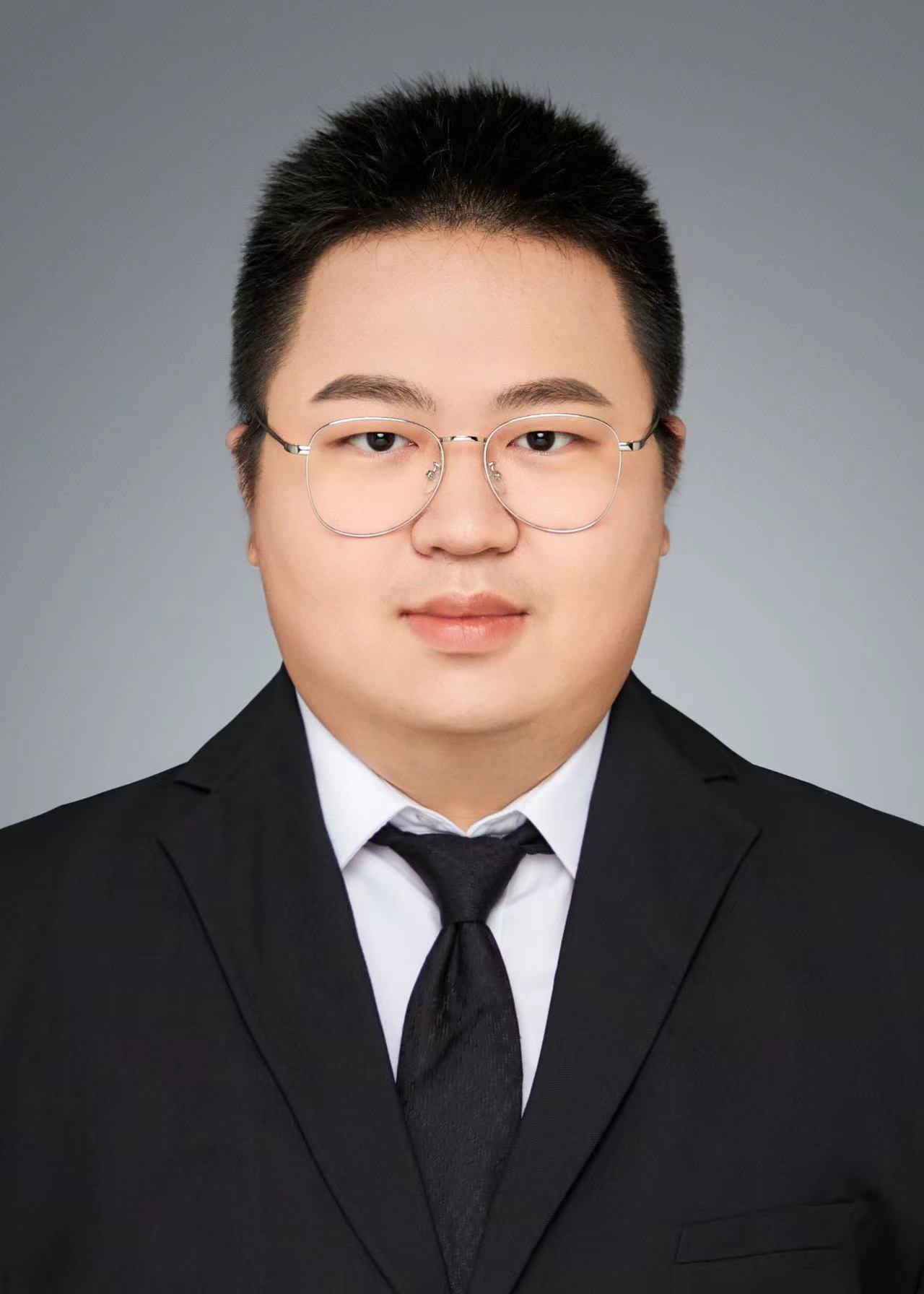}}]{Yiming Li} is currently a Research Professor in the School of Cyber Science and Technology at Zhejiang University. Before that, he received his Ph.D. degree with honors in Computer Science and Technology from Tsinghua University (2023) and his B.S. degree with honors in Mathematics from Ningbo University (2018). His research interests are in the domain of Trustworthy ML and Responsible AI, especially backdoor learning and AI copyright protection. His research has been published in multiple top-tier conferences and journals, such as ICLR, NeurIPS, and IEEE TIFS. He served as the Area Chair of ACM MM and Senior Program Committee Member of AAAI, and the reviewer of IEEE TPAMI, IEEE TIFS, IEEE TDSC, etc. His research has been featured by major media outlets, such as IEEE Spectrum. He was the recipient of the Best Paper Award at PAKDD and the Rising Star Award at WAIC.
\end{IEEEbiography}

\begin{IEEEbiography}[{\includegraphics[width=1in,height=1.2in,clip,keepaspectratio]{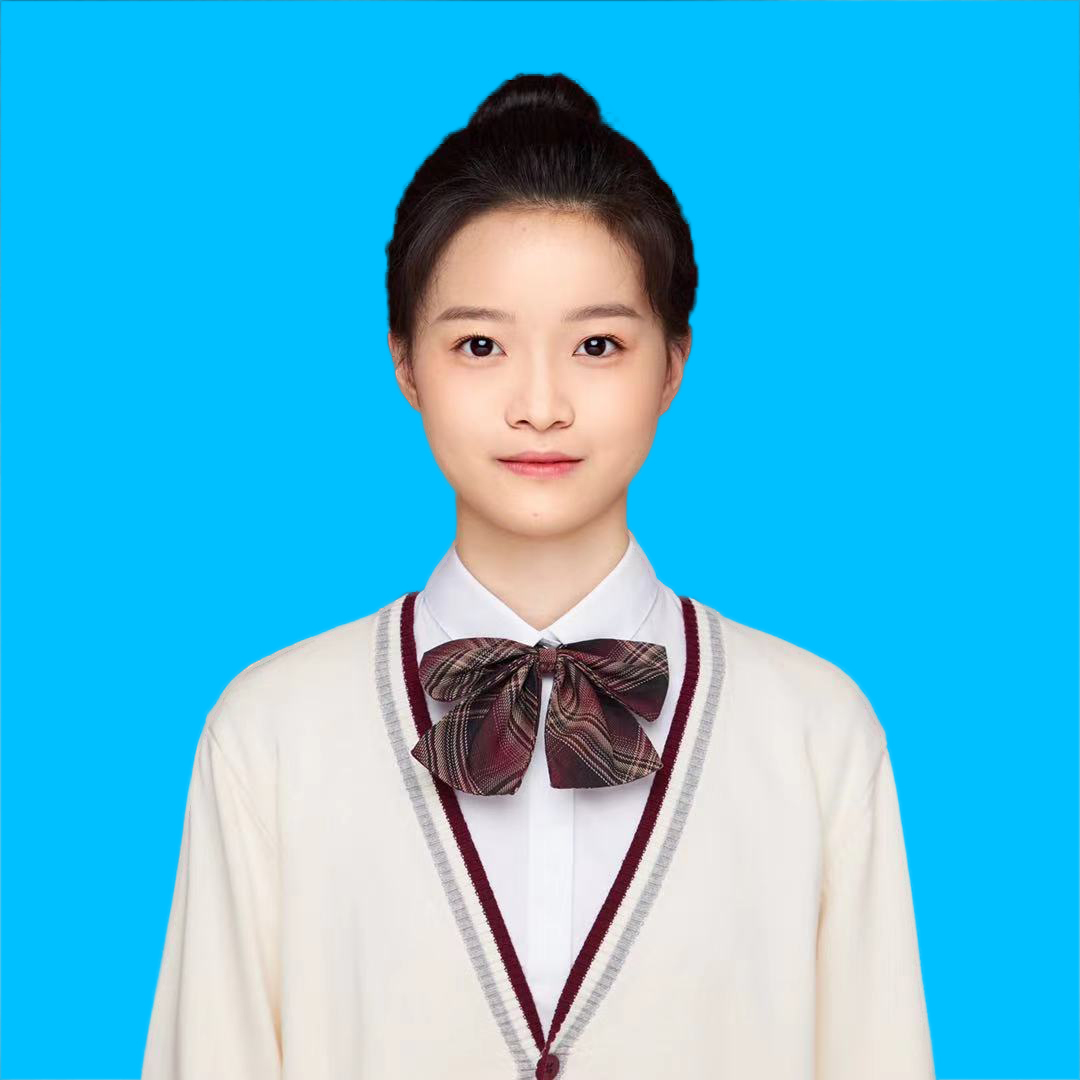}}]{Mengyuan Sun} is currently pursuing the B.E. at the School of Cyber Science and Engineering from Wuhan University, China. Her research interest is AI security.
\end{IEEEbiography}

\begin{IEEEbiography}[{\includegraphics[width=1in,height=1.25in,clip,keepaspectratio]{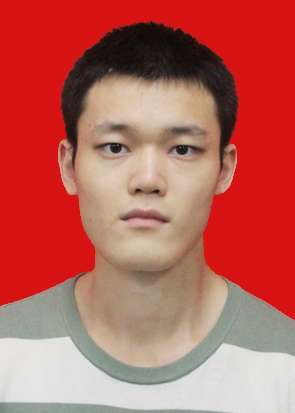}}]{Shuaike Li} is currently pursuing the B.E. at the School of Cyber Science and Engineering from Wuhan University, China. His research interest is AI security.
\end{IEEEbiography}

\begin{IEEEbiography}[{\includegraphics[width=1in,height=1.25in, clip,keepaspectratio]{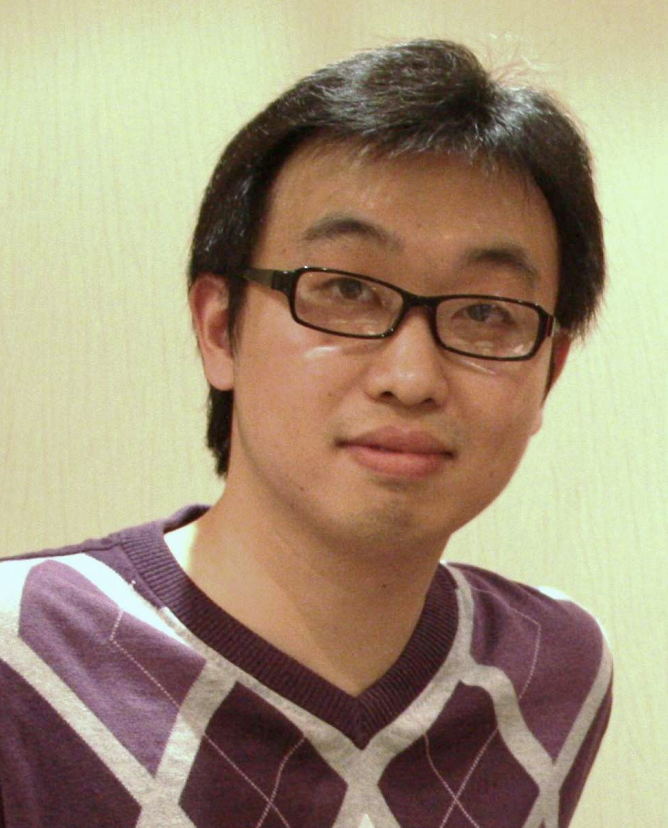}}]{Qian Wang} is a Professor in the School of Cyber Science and Engineering at Wuhan University, China. He was selected into the National High-level Young Talents Program of China, and listed among the World's Top 2\% Scientists by Stanford University. He also received the National Science Fund for Excellent Young Scholars of China in 2018. He has long been engaged in the research of cyberspace security, with focus on AI security, data outsourcing security and privacy, wireless systems security, and applied cryptography. He was a recipient of the 2018 IEEE TCSC Award for Excellence in Scalable Computing (early career researcher) and the 2016 IEEE ComSoc Asia-Pacific Outstanding Young Researcher Award. He has published 200+ papers, with 120+ publications in top-tier international conferences, including USENIX NSDI, ACM CCS, USENIX Security, NDSS, ACM MobiCom, ICML, etc., with 20000+ Google Scholar citations. He is also a co-recipient of 8 Best Paper and Best Student Paper Awards from prestigious conferences, including ICDCS, IEEE ICNP, etc. In 2021, his PhD student was selected under Huawei's  ``Top Minds'' Recruitment Program. He serves as Associate Editors for IEEE Transactions on Dependable and Secure Computing (TDSC) and IEEE Transactions on Information Forensics and Security (TIFS). He is a fellow of the IEEE, and a member of the ACM.
\end{IEEEbiography}

\begin{IEEEbiography}[{\includegraphics[width=1in,height=1.25in,clip,keepaspectratio]{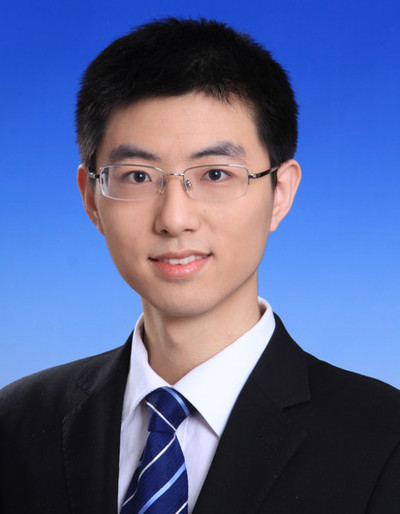}}]{Chen Chen} received his Ph.D. degree in computer science from Nanyang Technological University, Singapore, in 2024, his Master of Computer Science from the University of New South Wales, Australia, in 2018, and his Bachelor degree from the University of Science and technology Beijing, China, in 2012. His research interests lie in the area of AI safety, Knowledge Graphs and Large Language Models.
\end{IEEEbiography}

\vfill

\end{document}